\documentclass[a4paper,fleqn]{double-column}

\usepackage[authoryear,longnamesfirst]{natbib}

\def\tsc#1{\csdef{#1}{\textsc{\lowercase{#1}}\xspace}}
\tsc{WGM}
\tsc{QE}

\begin{document}
\let\WriteBookmarks\relax
\def\floatpagepagefraction{1}
\def\textpagefraction{.001}

\shorttitle{High-throughput 3D shape completion}    

\shortauthors{Blok et~al.}  

\title [mode = title]{High-throughput 3D shape completion of potato tubers on a harvester}  

\tnotemark[1] 

\tnotetext[1]{This study is partially funded by the Sarabetsu Village "Endowed Chair for Field Phenomics" project in Hokkaido, Japan. It is partially funded by the Deutsche Forschungsgemeinschaft (DFG, German Research Foundation) under Germany's Excellence Strategy, EXC-2070 -- 390732324 -- PhenoRob.} 

\author[1]{Pieter M. Blok}[type=editor,
      orcid=0000-0001-9535-5354]
\cormark[1]
\credit{Conceptualization, Methodology, Software, Data curation, Writing - original draft}

\author[2]{Federico Magistri}[type=editor,
      orcid=0000-0003-2815-5760]
\credit{Methodology, Software, Writing - review \& editing}

\author[2]{Cyrill Stachniss}[type=editor,
      orcid=0000-0003-1173-6972]
\credit{Methodology, Software, Writing - review \& editing}

\author[1]{Haozhou Wang}[type=editor,
        orcid=0000-0001-6135-402X]
\credit{Data curation, Resources, Software, Writing - review \& editing}

\author[1]{James Burridge}[type=editor,
        orcid=0000-0002-2194-9894]
\credit{Conceptualization, Methodology, Data curation, Supervision, Writing - review \& editing}

\author[1]{Wei Guo}[type=editor,
        orcid=0000-0002-3017-5464]
\credit{Conceptualization, Methodology, Funding acquisition, Project administration, Supervision, Writing - review \& editing}

\affiliation[1]{organization={Graduate School of Agricultural and Life Sciences, The University of Tokyo},
            addressline={1-1-1 Midori-cho}, 
            city={Nishitokyo-city},
            postcode={188-0002}, 
            state={Tokyo},
            country={Japan}}
            
\affiliation[2]{organization={Center for Robotics, University of Bonn},
            addressline={Nussallee 15}, 
            city={Bonn},
            postcode={53115}, 
            state={North Rhine-Westphalia},
            country={Germany}}

\cortext[1]{Corresponding author: pieter.blok@fieldphenomics.com (P.M. Blok).}


\begin{abstract}
Potato yield is an important metric for farmers to further optimize their cultivation practices. Potato yield can be estimated on a harvester using an RGB-D camera that can estimate the three-dimensional (3D) volume of individual potato tubers. A challenge, however, is that the 3D shape derived from RGB-D images is only partially completed, underestimating the actual volume. To address this issue, we developed a 3D shape completion network, called CoRe++, which can complete the 3D shape from RGB-D images. CoRe++ is a deep learning network that consists of a convolutional encoder and a decoder. The encoder compresses RGB-D images into latent vectors that are used by the decoder to complete the 3D shape using the deep signed distance field network (DeepSDF). To evaluate our CoRe++ network, we collected partial and complete 3D point clouds of 339 potato tubers on an operational harvester in Japan. On the 1425 RGB-D images in the test set (representing 51 unique potato tubers), our network achieved a completion accuracy of 2.8 mm on average. For volumetric estimation, the root mean squared error (RMSE) was 22.6 ml, and this was better than the RMSE of the linear regression (31.1 ml) and the base model (36.9 ml). We found that the RMSE can be further reduced to 18.2 ml when performing the 3D shape completion in the center of the RGB-D image. With an average 3D shape completion time of 10 milliseconds per tuber, we can conclude that CoRe++ is both fast and accurate enough to be implemented on an operational harvester for high-throughput potato yield estimation. CoRe++'s high-throughput and accurate processing allows it to be applied to other tuber, fruit and vegetable crops, thereby enabling versatile, accurate and real-time yield monitoring in precision agriculture. Our code, network weights and dataset are publicly available at \url{https://github.com/UTokyo-FieldPhenomics-Lab/corepp.git}. \nocite{*}
\end{abstract}



\maketitle

\section{Introduction}
\label{introduction}
Potatoes (\textit{Solanum tuberosum}) are important for global food security \citep{zhang2017} and are of considerable economic importance. In 2021, potato production contributed \$100 billion to the US economy \citep{knudson2023}, and €19.4 billion to the EU economy in 2023 \citep{eurostat2024}. Potatoes have relatively high nutrient requirements and the quantity and timing of nutrient application impacts tuber yield \citep{ruark2014, love2005}. Excess nutrient application represents wasted money and can contribute to waterway eutrophication \citep{davenport2005}. While fields are typically uniformly fertilized, variable rate precision fertilization offers the potential to reduce expenses while optimizing tuber yield and reducing environmental impact \citep{ahmad2023}. To enable precision fertilization and to further optimize global resource management, farmers must have data on field yields. Such field yield data can be estimated with above-ground biomass measurements \citep{liu20241, liu2024} or close-range monitoring systems installed on the harvester. Such monitoring systems can automatically estimate key yield parameters such as tuber number, size, volume, and weight. Commonly used monitoring systems use load cells attached to the harvesters' conveyor belt to measure the mass of the harvested produce in real time \citep{zamani2014, kabir2018}. Such weighing systems are easy to use and maintain, but disadvantages of these systems is that they only capture gross yield and that they include tare (such as stones, soil and crop residue) in the weight measurement. Therefore, there is a need for a tare-free and individualized monitoring of tubers, which allows farmers to learn from their fertilizer management decisions and enable them to collect detailed information on marketable yields for better sales decisions. 

Tare-free and individualized monitoring of potato tubers can be accomplished by a camera system and computer vision software. Currently, at least 14 of such camera systems have been presented in scientific literature: \citet{noordam2000, hofstee2003, elmasry2012, razmjooy2012, lee2018, long2018, si2018, su2018, pandey2019, cai2020, lee2020, dolata2021, huynh2022, jang2023}. Most of these systems (11 of the 14) used a single red, green and blue (RGB) color camera to detect the potato tubers and then estimate their yield parameters using offline calibrated pixel-to-world conversion factors. While this method proved effective in calibrated laboratory setups, it was also found to have limited accuracy for potatoes that are occluded. Occlusion is a common phenomena on the conveyor belt of an operational harvester.

More recent studies have focused on extending the traditional RGB camera with additional three-dimensional (3D) vision. With 3D vision, it is possible to create partial or complete 3D shapes of potato tubers, allowing for better estimation of tuber yield under the challenging conditions of occlusion. \citet{cai2020} extended an RGB camera with a laser line triangulation method to capture the complete 3D shape of potato tubers. Although their system enabled very accurate volumetric estimates of potato tubers in a laboratory setup, the image acquisition method was unfortunately too time-consuming to be applied on an operational harvester. RGB cameras with embedded depth sensing abilities, known as RGB-D cameras, allow much higher throughput than laser triangulation methods. Although being more susceptible to dust and motion blur, RGB-D cameras are cheaper and easier to integrate into existing machines than laser triangulation methods. RGB-D cameras, such as the ones used by \citet{long2018} and \citet{su2018}, are therefore more promising for use on an operational harvester. 

A current limitation with RGB-D cameras is that they can only produce a partial 3D shape of the potato tuber, which can lead to an underestimation of the actual size, volume or weight. Combining multiple RGB-D cameras can potentially reduce the effect of this problem, but this will not fully solve it and make the overall system complex and more expensive. Therefore, it is more desirable and economically feasible to use a single RGB-D camera, and then estimate the complete 3D shape with dedicated software. There are numerous examples in the scientific literature of using 3D shape completion software to estimate complete shapes from partial shapes. Most of the current shape completion methods use deep learning techniques based on multi-layered perceptrons, graph-based convolutional neural networks and encoder-decoder networks to complete the 3D shape from partially completed point clouds \citep{fei2022}.
Also in the agricultural domain, where deep learning applications are widespread for disease detection, yield prediction and machine automation, there are studies on 3D shape completion: one on the completion of plant leaves \citep{chen2023}, one on the completion of trees \citep{xu2023}, and five on the completion of fruits \citep{ge2020, magistri2022, magistri2024, marangoz2020, pan2023}. 

When assessing the applicability of 3D shape completion methods for potato yield estimation, it is crucial to consider their processing speed. Ideally, the shape completion method should be fast enough to process all the potato tubers that move over the conveyor belt during harvest, because this will provide the most complete and accurate yield estimate possible. Since roughly more than one million potato tubers are harvested per hectare, the 3D shape completion method must complete processing in a matter of milliseconds. Of the studies mentioned above, only the method of \citet{magistri2022} has the potential to do so. Magistri's method, called CoRe (Completion and Reconstruction), can complete 3D shapes in 4 milliseconds (tested on sweet peppers and strawberries). The novelty of CoRe is the addition of a convolutional encoder to the deep signed distance function decoder (DeepSDF, \citealp{park2019}). CoRe's convolutional encoder compresses RGB-D images into a latent vector, which is a compact representation of the shape's geometry. The latent vector from the encoder is then used along with the 3D coordinates of the point cloud as input for DeepSDF to reconstruct the complete 3D shape. Because the latent vector is predicted by the convolutional encoder, there is no need for DeepSDF's original time-consuming latent vector optimization, making the entire 3D shape completion method significantly faster.

Although the work by \citet{magistri2022} shows the potential for high-throughput 3D shape completion, there are still unanswered research questions. First, what is the optimal size of the latent vector for implicitly learning the shape of the potato tubers? Second, given that the potato tubers are transported on the conveyor belt and thus move from bottom to top in the image, what is the best image location to perform the 3D shape completion? Third, and most important, is the 3D shape completion method able to quickly and accurately estimate the volume of fast moving potato tubers on an operational harvester?

The main contribution of this paper is the extension of the work by \citet{magistri2022} to answer these research questions. We have conducted a systematic study and analysis on how to optimize the 3D shape completion method. Our optimization has led to several improvements to CoRe's original convolutional encoder: an improved data preprocessing, geometric and color data augmentations, an updated loss function, an updated neural network architecture, and a graphical processing unit (GPU)-based 3D mesh generation that improves and speeds up the 3D shape completion for precision agriculture applications. 

This paper presents the research and development of a 3D shape completion network for estimating the volume of individual potato tubers on an operational harvester.  Our research novelties are four-fold. First, we optimized CoRe's original convolutional encoder for faster and more accurate 3D shape completion of potato tubers on an operational harvester. Second, we performed a systematic analysis of the effect of the latent size, image analysis region, potato size, and potato cultivar on the performance of the 3D shape completion of potato tubers. Third, we conducted two ablation studies on the impact of our new additions to the overall performance. Fourth, we publicly released our 3D dataset with partially and fully completed point clouds of potato tubers collected on an operational harvester in dirty and cluttered circumstances. Although this study focused on potato tubers, our shape completion method can also be applied to other crops that suffer from object occlusion, such as fruits, berries, peppers, tomatoes and other root crops. Our method can thus serve the broader purpose of better handling partial visibility in agricultural applications requiring high throughput, such as robotic harvesting, pick-and-place, path planning, and obstacle detection and avoidance.

\section{Materials and methods}
\label{materials_methods} 

\subsection{Dataset}
\label{data_collection}

\subsubsection{Imaging system}
\begin{figure*}[hbt!]
  \centering
  \subfloat[] {\includegraphics[width=1.0\textwidth]{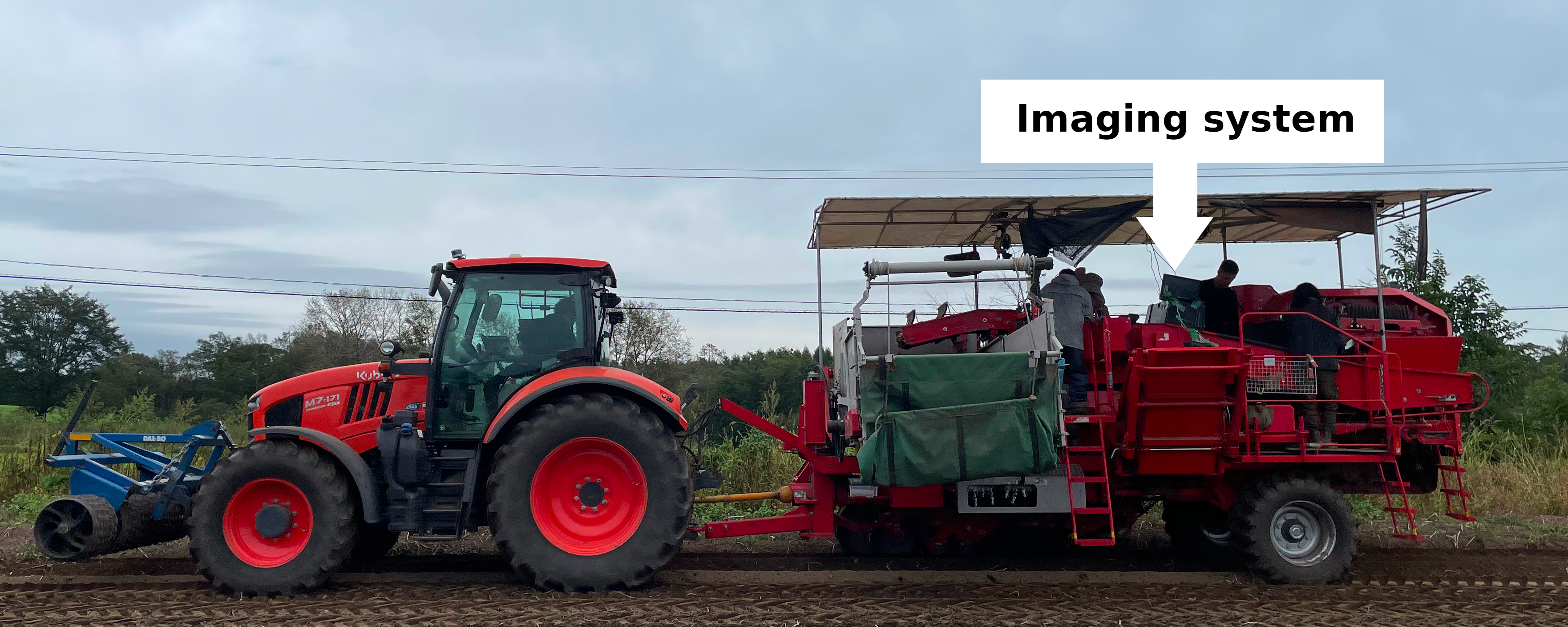}\label{fig:harvester}}
  \hfill
  \subfloat[] {\includegraphics[width=0.62\textwidth]{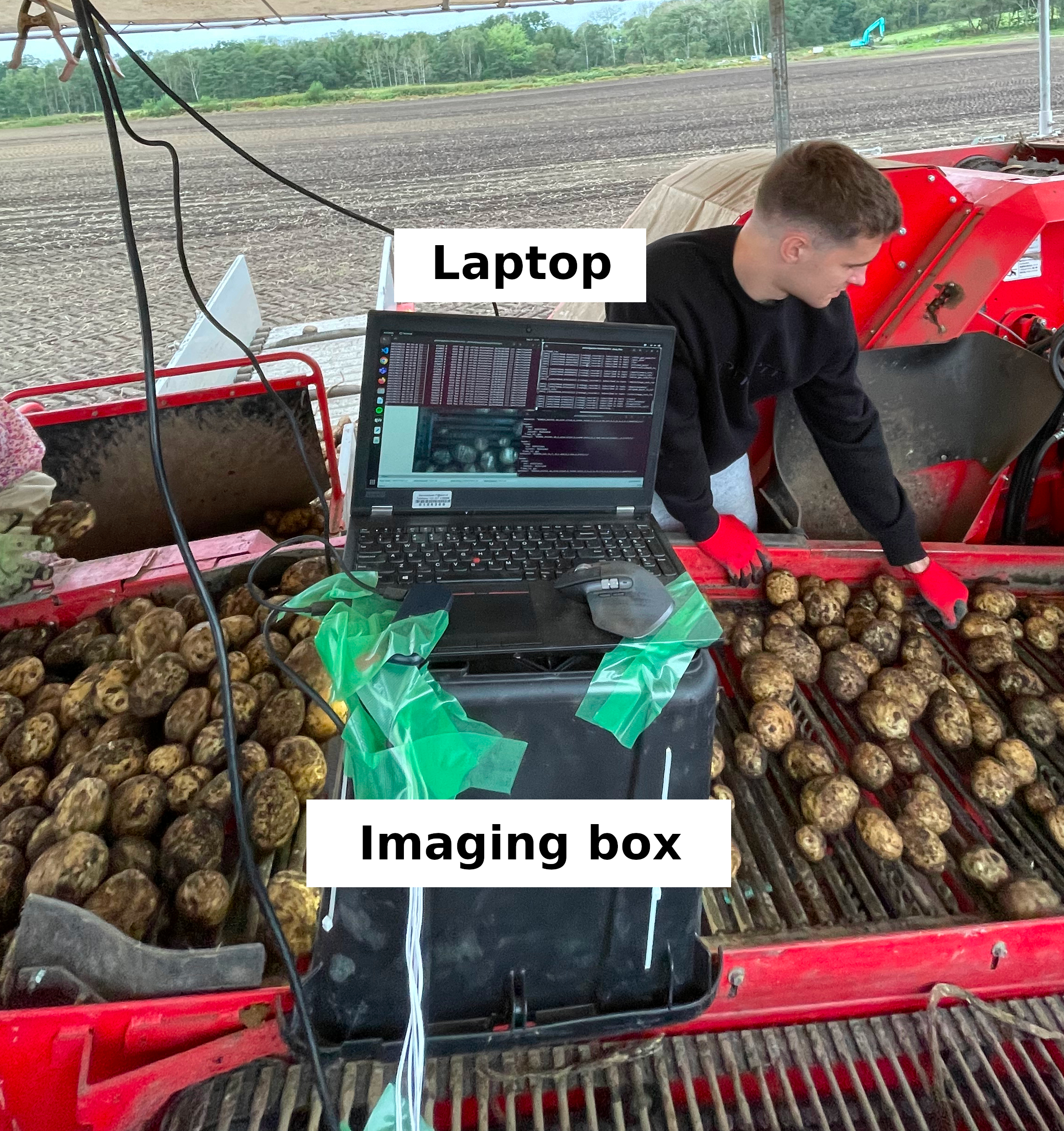}\label{fig:box_on_harvester}}
  \hfill
  \subfloat[] {\includegraphics[width=0.3638\textwidth]{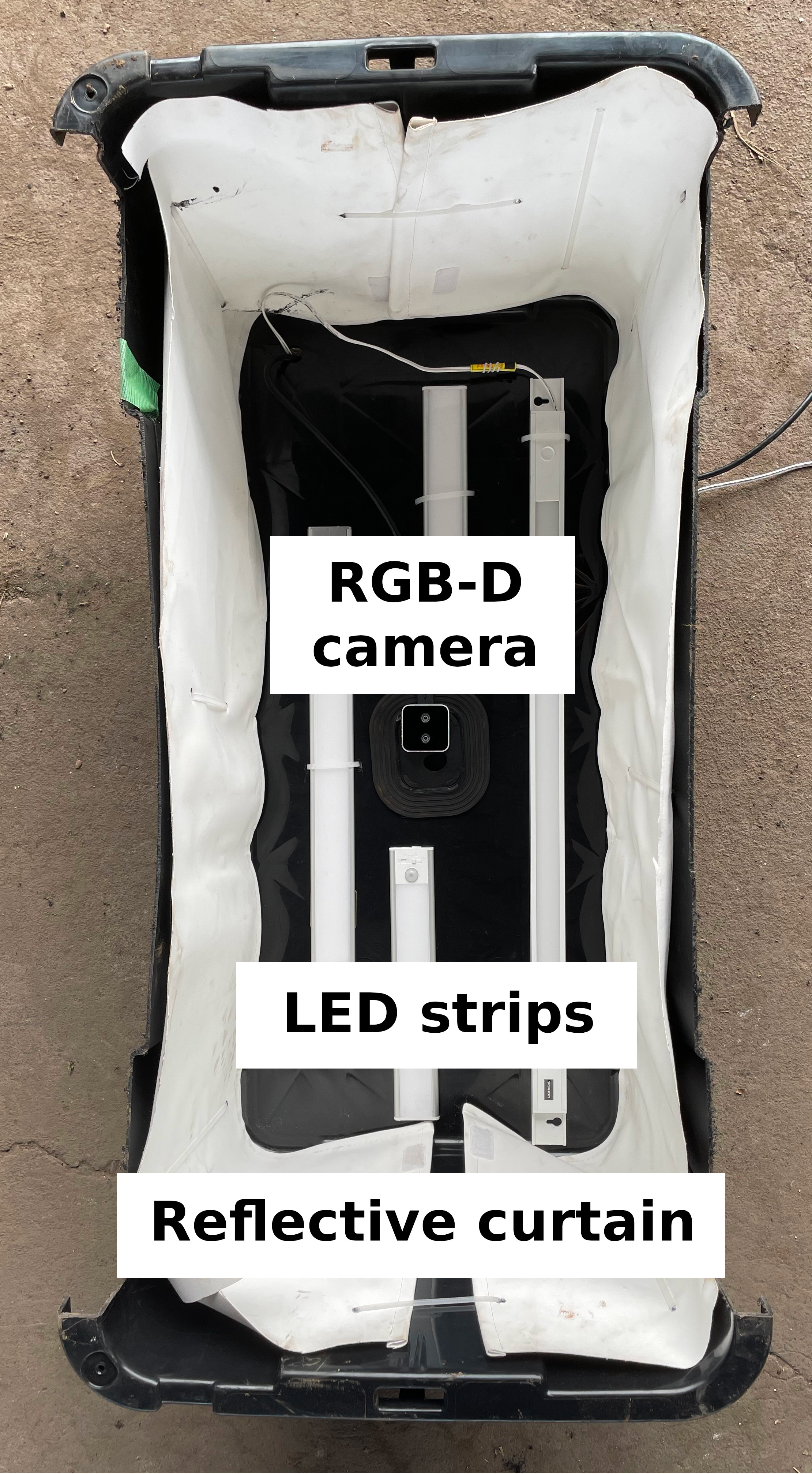}\label{fig:box_inside}}
  \caption{(a) and (b) give overviews of the imaging system installed on a potato harvester in Sarabetsu, Japan. (c) Inside the imaging box, an RGB-D camera was installed, together with four LED strips that provided the necessary illumination inside the box. The sides of box were covered with a reflective curtain to generate diffuse lighting conditions.}
  \label{fig:data_collection_system}
\end{figure*}

We installed an imaging system above the conveyor belt of a single-row potato harvester (Toyonoki Top-1, Figure~\ref{fig:harvester}). The imaging system was installed in a black plastic box with dimensions of 85 cm (width), 45 cm (depth) and 39 cm (height) (Figure~\ref{fig:box_on_harvester}). Inside the plastic box, four light emitting diode (LED) strips were installed to provide light with a color temperature of 6000K (Figure~\ref{fig:box_inside}). Our enclosed setup not only ensured uniform lighting conditions, which improved the RGB-D image quality, but also ensured that our system could be used at night. Reflective curtains on the sides of the plastic box helped to diffuse the light. In the center of the box, an RGB-D camera (Intel Realsense D405) was installed that captured images of the conveyor belt with a top-view perspective. The distance between the RGB-D camera and the conveyor belt was approximately 0.33 m. At this distance, the camera's field of view was 0.64 m (width) by 0.39 m (height). The RGB-D camera was connected to a Lenovo Thinkpad P53 laptop with an Intel Xeon E-2276M 2.8 GHz CPU with 64GB RAM and a Nvidia Quadro RTX 5000 GPU with 16GB memory. On the laptop, the Robot Operating System 2 (ROS2, version: Humble Hawksbill) was used to collect color and depth images with 30 frames per second. The exposure time of the RGB-D camera was set to 5.0 ms, allowing us to capture the images without motion blur. The  images were stored in ROS2 bag files. 

\subsubsection{Image collection on the harvester}
RGB-D images were collected from 12 rows in a potato field in Sarabetsu, Japan (latitude: 42.610316, longitude: 143.156753). The row spacing in this field was 0.75 m and the potato cultivar was Sayaka. For each row, a separate ROS2 bag file was recorded and stored on the laptop. To increase diversity in potato sizes and shapes, we also collected data by running the harvester in the barn of the farm and then dumping boxes of two different potato cultivars on the conveyor belt. Although this mimicked the density of potato tubers on the conveyor belt in the field, there was no tare in the boxes, making the detection task easier. The barn experiment was conducted with six boxes of potatoes, each weighing between eight and ten kilograms. The six boxes were divided into three boxes with potatoes from the Kitahime cultivar and three boxes from the Corolle cultivar. A separate ROS2 bag file was recorded for each box and stored on the laptop.

\begin{figure*}[hbt!]
  \centering
  \subfloat[] {\includegraphics[width=0.325\textwidth]{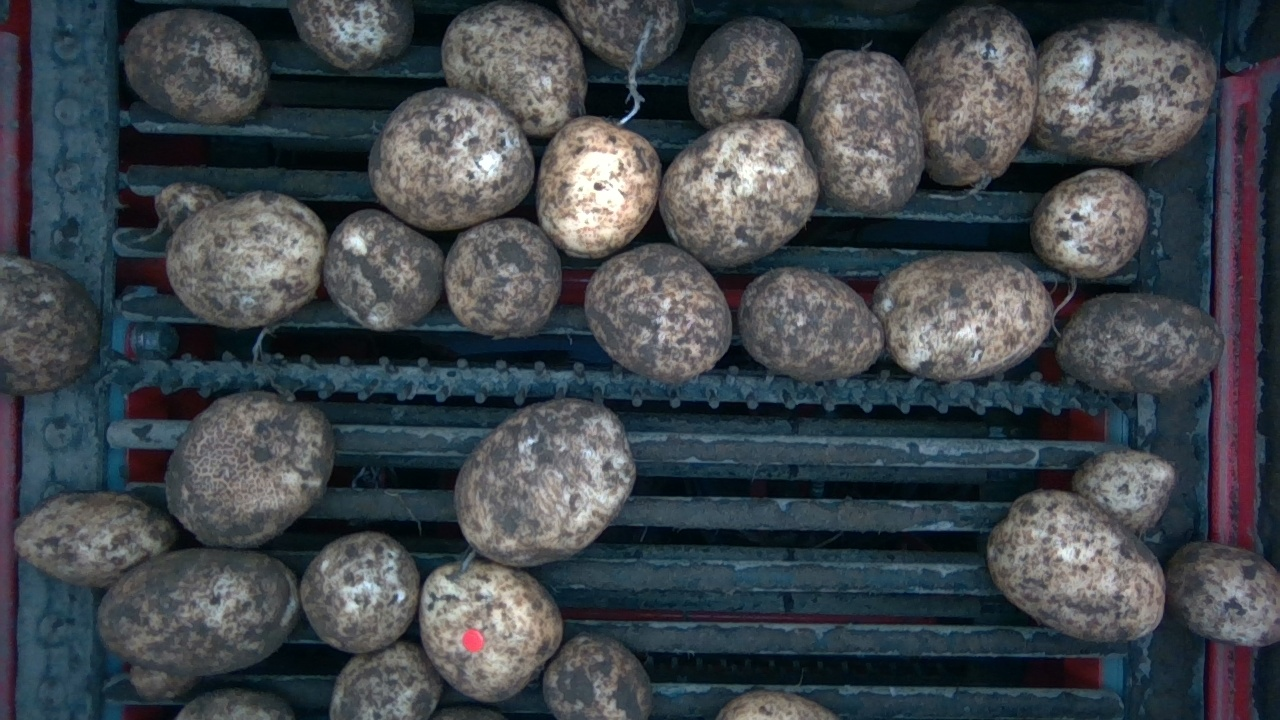}\label{fig:frame_01}}
  \hfill
  \subfloat[] {\includegraphics[width=0.325\textwidth]{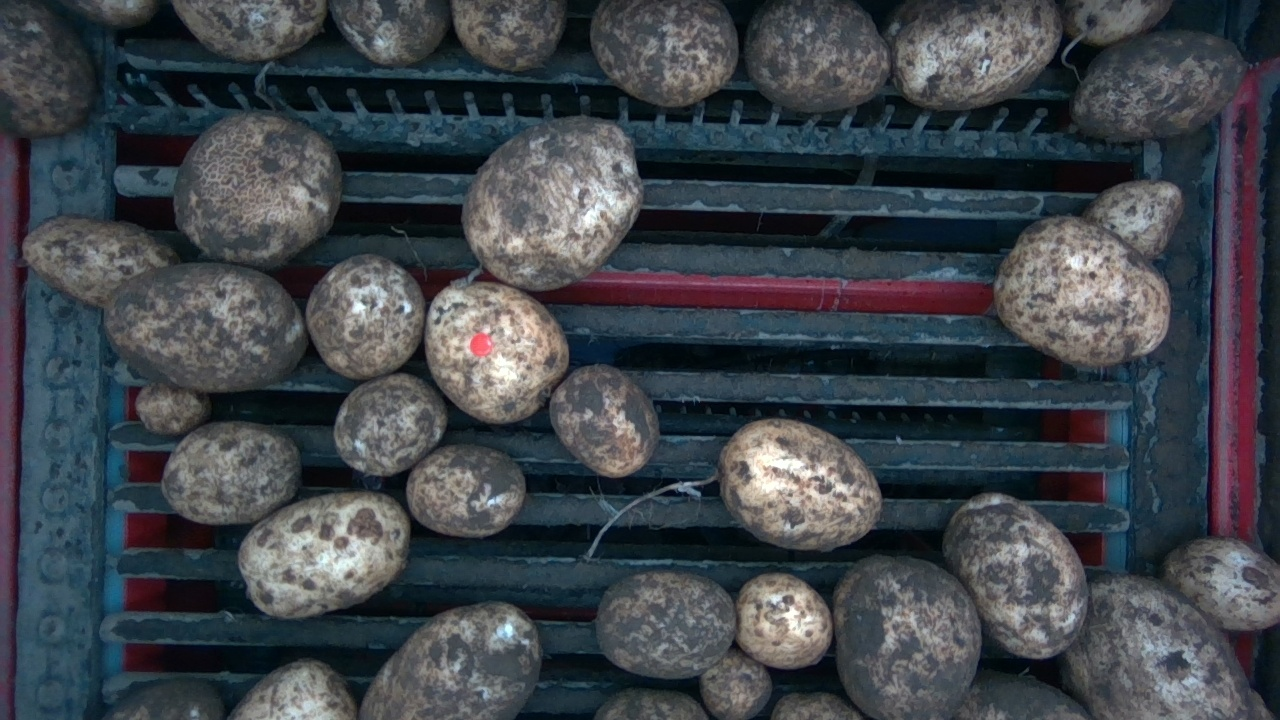}\label{fig:frame_02}}
  \hfill
  \subfloat[] {\includegraphics[width=0.325\textwidth]{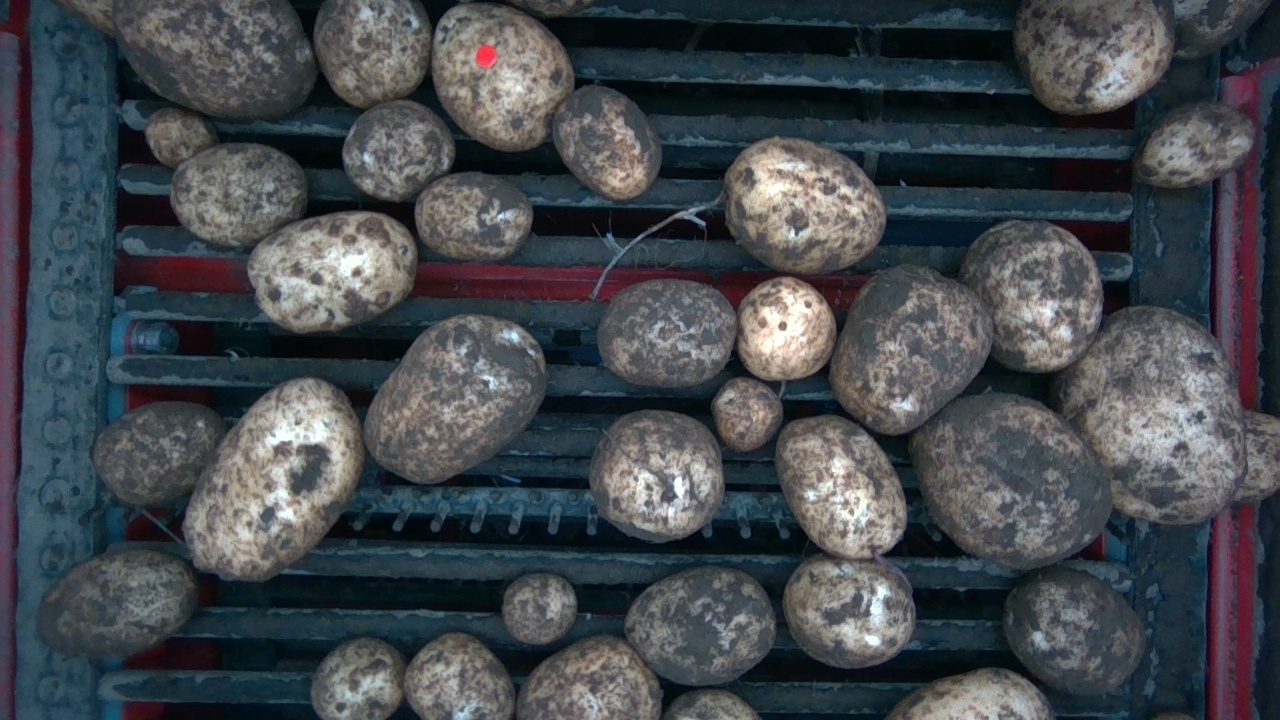}\label{fig:frame_03}}
  \caption{Our potato collection method involved marking potato tubers with a colored thumbtack so that the tuber could be easily identified in the image and easily collected after image acquisition. (a), (b), and (c) show a tuber marked with a red thumbtack while it moved over the conveyor belt.}
  \label{fig:frames}
\end{figure*}

While running the harvester in the field or in the barn, we collected images of 339 potato tubers of different sizes and shapes to test our 3D shape completion method. Our collection method consisted of the following procedure: one person standing in front of the imaging box randomly selected a potato tuber and then inserted a colored thumbtack into the potato. The thumbtack was inserted into the potato such that it was visible in the RGB image when the tuber passed the image acquisition region of the RGB-D camera. This procedure was repeated for all 339 potato tubers. Given the speed of the conveyor belt, we were able to capture between 20 and 30 images for each marked potato tuber when it moved under the camera along with the conveyor belt (Figure \ref{fig:frames}). After the image acquisition, another person standing behind the imaging box collected the marked potato by hand. The marked potato was then placed in a bucket that was later brought to the barn for 3D reconstruction. 

\subsubsection{3D reconstruction of potato tubers as ground truth}
\label{3d_reconstruction}
\begin{figure*}[hbt!]
  \centering
    \includegraphics[width=0.95\textwidth]{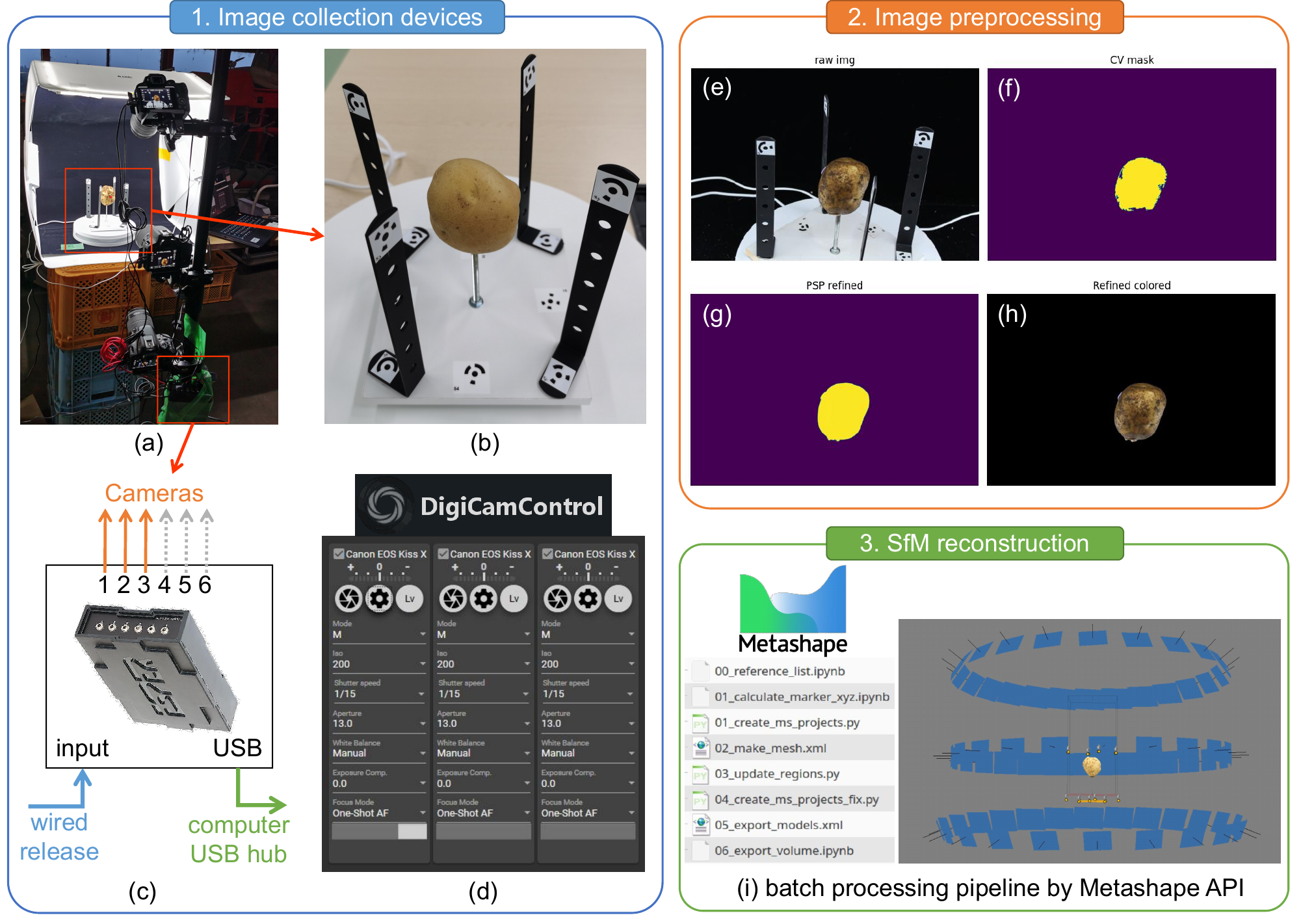}
    \caption{The workflow of our 3D reconstruction included three steps: (1) image collection, (2) image preprocessing, (3) 3D reconstruction with Structure-from-Motion (SfM).}
    \label{fig:sfm_devices}
\end{figure*}

To obtain the complete 3D shape of the collected potato tubers as ground truth, we have set up a 3D reconstruction system in the barn (Figure~\ref{fig:sfm_devices}a). This system consisted of three Canon X7 DSLR cameras, a turntable (Foldio 360, OrangeMonkie Inc.), four auto-detectable marker stands (Fig.~\ref{fig:sfm_devices}b), and a photo studio with LED illumination. 

Prior to taking the images, each collected potato tuber was pierced with a narrow threaded bolt attached to a white wooden board (Figure~\ref{fig:sfm_devices}b). This piercing allowed the tuber to be photographed from three camera perspectives while being held off the ground, allowing almost the entire longitudinal section of the tuber to be photographed at once. To capture all sides of the tuber, the turntable was set to rotate 15 degrees per interval and then stop for two seconds to give the three cameras time to acquire the images. The image acquisition was accomplished with a commercial shutter controller (Esper TriggerBox) that released an electronic trigger to the three cameras at the same time. For each camera, 24 images were recorded per complete rotation of the turntable. The resulting 72 images were directly transferred to the connected laptop computer to store them for 3D reconstruction (Figure~\ref{fig:sfm_devices}c). The camera parameter settings, image transferring, and image renaming were accomplished with DigiCamControl software, of which its details are shown in Figure~\ref{fig:sfm_devices}d.

\begin{figure*}[hbt!]
  \centering
  \subfloat[] {\includegraphics[width=0.2\textwidth]{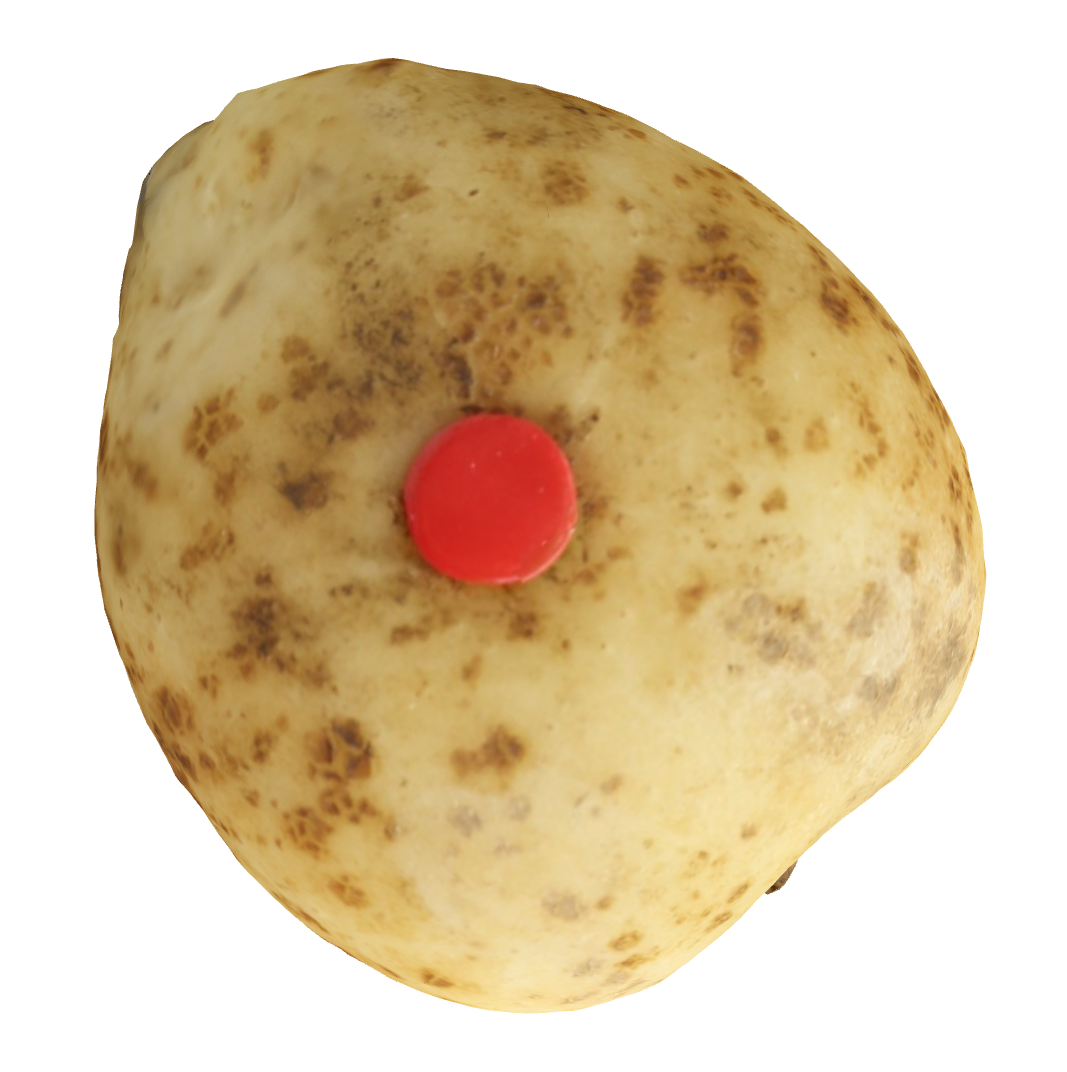}\label{fig:3d_frame_01}}
  \hfill
  \subfloat[] {\includegraphics[width=0.2\textwidth]{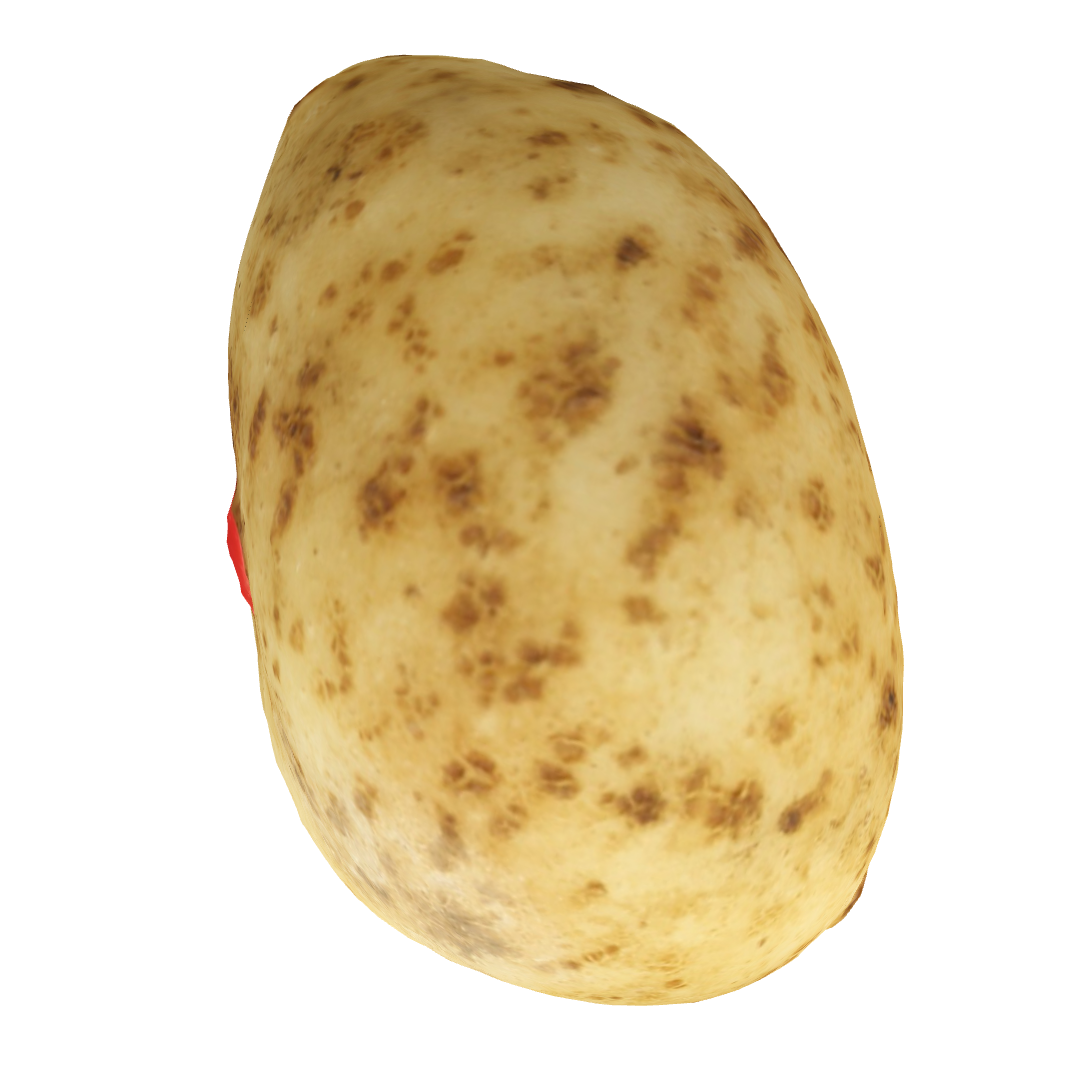}\label{fig:3d_frame_02}}
  \hfill
  \subfloat[] {\includegraphics[width=0.2\textwidth]{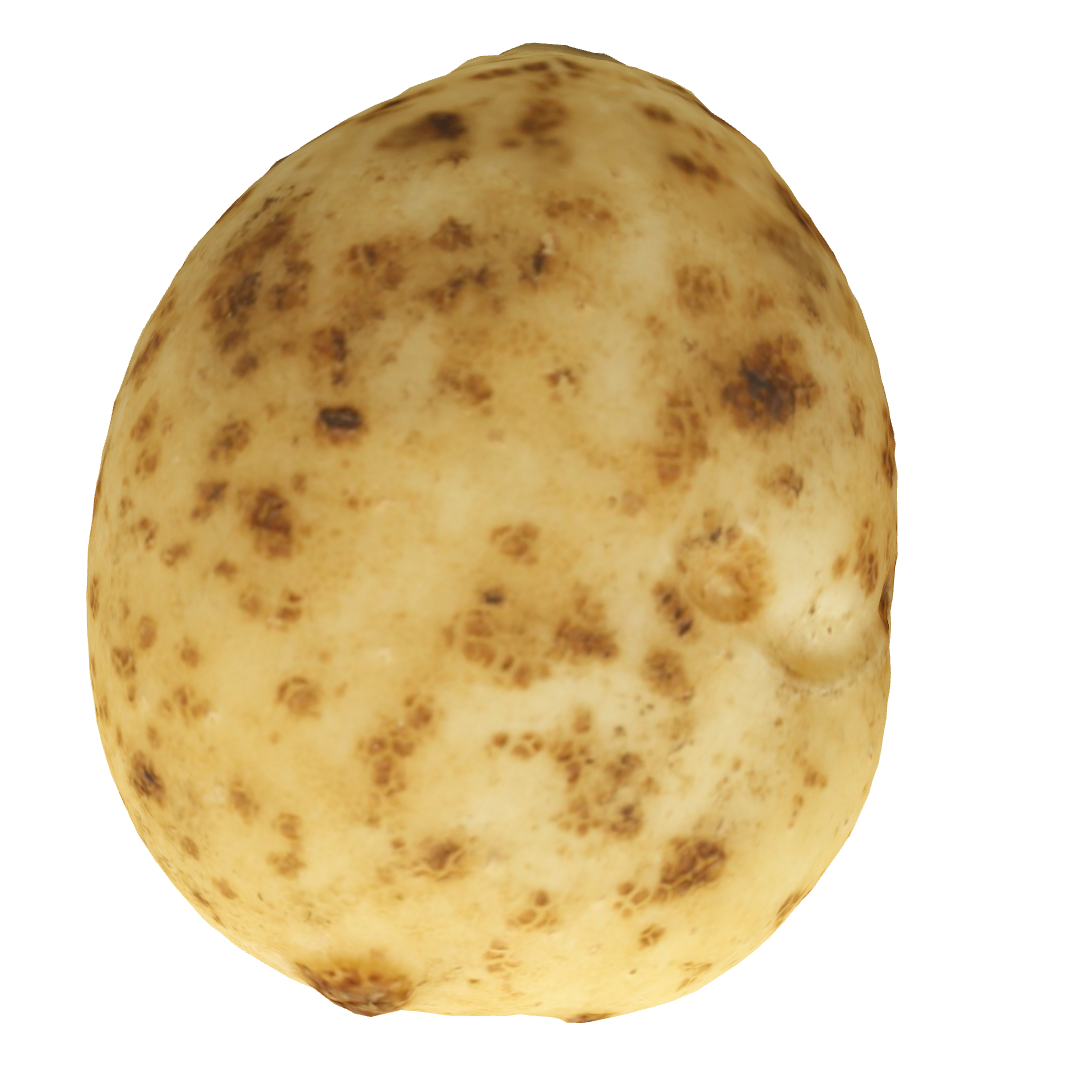}\label{fig:3d_frame_03}}
  \hfill
  \subfloat[] {\includegraphics[width=0.2\textwidth]{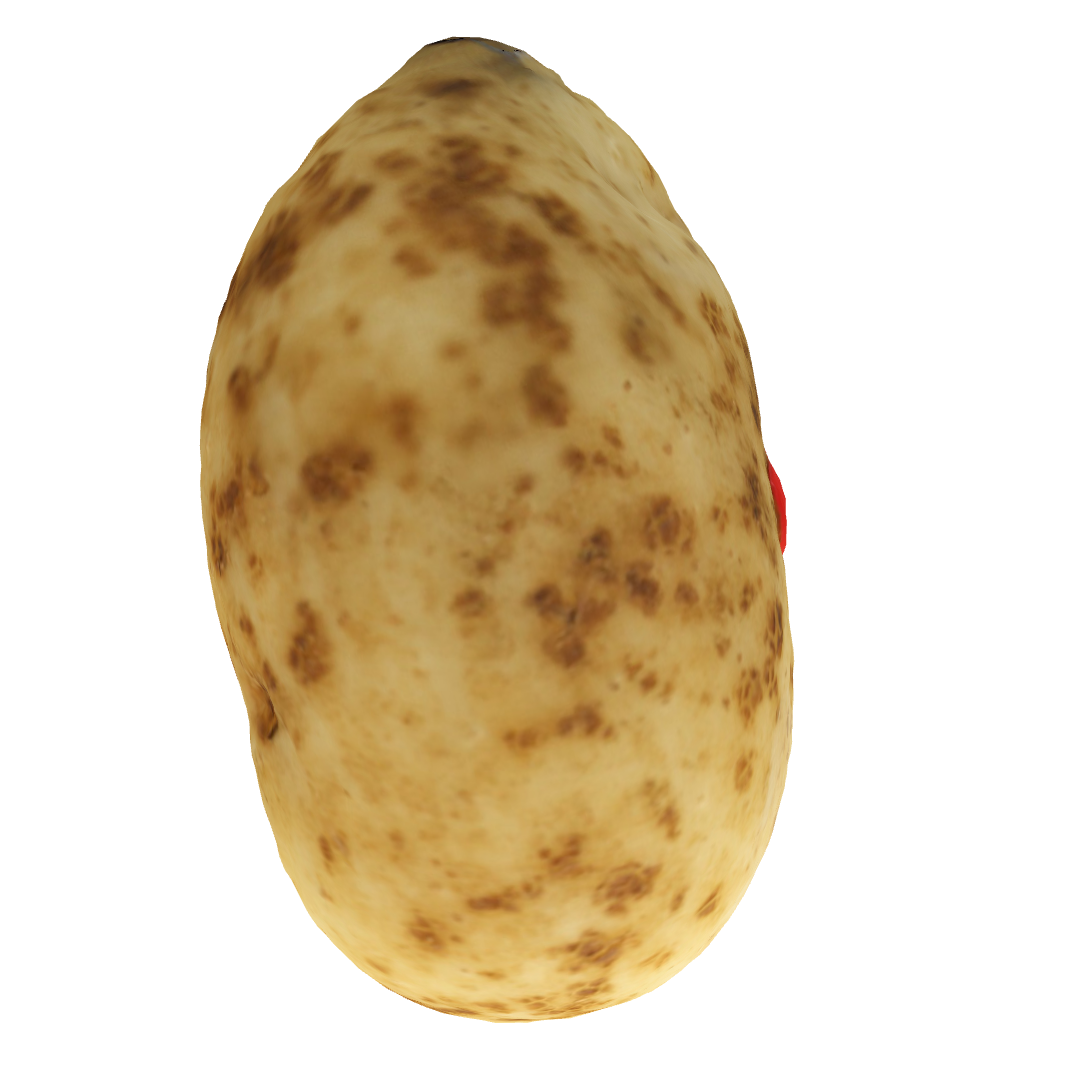}\label{fig:3d_frame_04}}
  \caption{3D colored mesh of a potato tuber produced by our 3D reconstruction pipeline. (a) front view, (b) right side view, (c) back view, (d) left side view.}
  \label{fig:3d_frames}
\end{figure*}

After the image acquisition, the images were processed to extract the mask region of the potato tuber. Due to soil residues and light reflections on the tuber surfaces (Figure~\ref{fig:sfm_devices}e), which were similar to the black and white background, we had to use a combination of color-based filtering and a deep learning-based segmentation refinement. Given that potato tubers are predominantly yellow (Figure~\ref{fig:3d_frames}), we chose to use the CIELAB color space, because yellow in CIELAB is represented with high positive values on the b* axis and near-zero values on the a* axis, allowing for straightforward filtering without interference from other colors or lighting conditions (we used b* $\geq$ 15 for filtering, see the result in Figure~\ref{fig:sfm_devices}f). By using CIELAB filtering as pre-segmentation, it became easier for the deep learning network to refine the mask (Figure~\ref{fig:sfm_devices}g). We used the CascadePSP \citep{cheng2020} deep learning network due to its multi-stage refinement approach, which enabled better segmentation quality compared to other segmentation networks like U-Net \citep{ronneberger2015} and Mask R-CNN \citep{he2017}.

After generating the masks of all tubers, we implemented an automatic batch processing workflow based on the Metashape API for 3D reconstruction (Figure~\ref{fig:sfm_devices}i). First, the images and corresponding masks of each potato tuber were grouped for each camera and its corresponding view angles. Then, the control point markers in the images were automatically detected, the scale bars were automatically imported, and world coordinates were automatically assigned to each potato tuber. Afterwards, the images of the different cameras were aligned, and the corresponding key points were generated. These aligned images served as input for the 3D reconstruction, which resulted the colored 3D mesh models for all 339 potato tubers in our dataset. After 3D reconstruction, small distortions, such as 3D meshes belonging to the white wooden board were removed. Then, the filtered meshes were double-checked by another researcher in CloudCompare who removed small disconnected components if necessary. Figure~\ref{fig:3d_frames} illustrates a filtered 3D mesh of a potato tuber with colored textures. 

\subsubsection{Dataset splits and volume}
\label{dataset_splits}
\begin{figure*}[hbt!]
  \centering
    \includegraphics[width=1\textwidth]{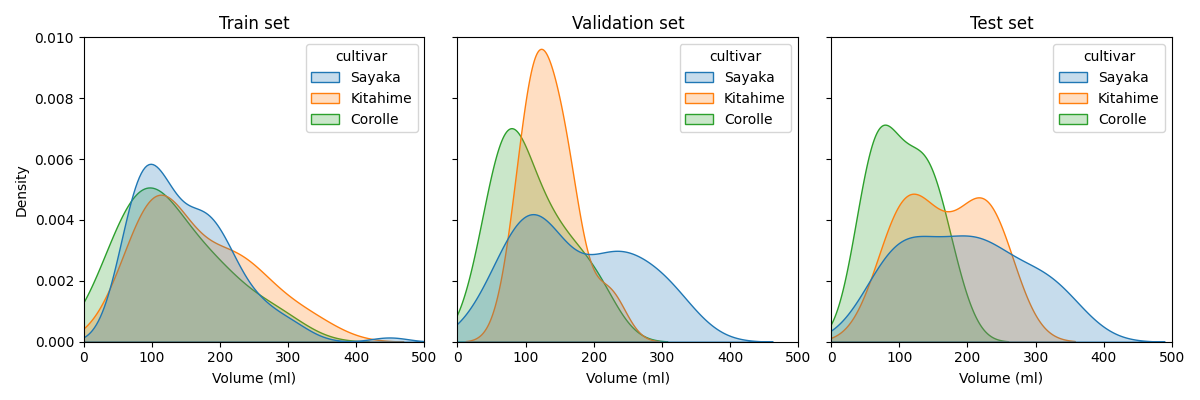}
    \caption{Kernel density estimate plots for visualizing the volumetric distribution by potato cultivar in the train, validation and test set.}
    \label{fig:histogram}
\end{figure*}

After the 3D reconstruction of the 339 collected potato tubers, we split the dataset into a train, validation and test set. The split was made in such a way that the three sets contained a representative portion of different sizes, shapes and cultivars (Figure~\ref{fig:histogram}). Our split ratio was 70\% for the train set (237 potato tubers), 15\% for the validation set (51 potato tubers) and 15\% for the independent test set (51 potato tubers). 

Because each potato tuber was photographed 20 to 30 times on the conveyor belt of the harvester, the total number of RGB-D images for training the encoder was 6794. The number of validation images was 1439 and the number of independent test images was 1425.

The average data volume per potato was 58.5 MB. This consisted of an average of 55.1 MB for the RGB-D image frames and an average of 3.5 MB for the 3D mesh.

\subsection{3D shape completion network}

\subsubsection{Encoder and decoder architecture}
\label{core_architecture}
\begin{figure*}[hbt!]
  \centering
    \includegraphics[width=1\textwidth]{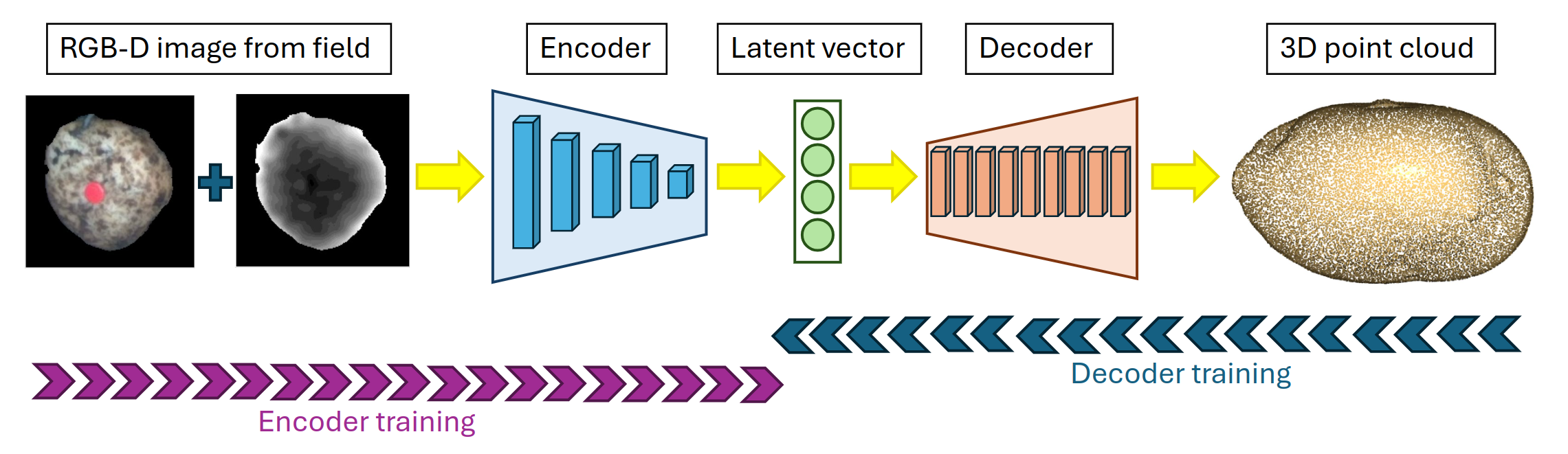}
    \caption{Schematic representation of our encoder-decoder network for 3D shape completion of potato tubers. The encoder compresses RGB-D images into a latent vector, which is processed by the DeepSDF decoder to reconstruct the complete 3D point cloud.}
    \label{fig:network}
\end{figure*}

Our 3D shape completion network was based on the CoRe network presented in \citet{magistri2022}. Given the many updates made to CoRe during our research, our 3D shape completion network was renamed to CoRe++. Figure \ref{fig:network} gives a schematic overview of CoRe++. CoRe++ consisted of a convolutional encoder and a decoder. This encoder-decoder network was chosen over alternatives like 3D convolutions or voxel-based networks due to its small network architecture, thereby enabling high-throughput processing. The encoder was a small neural network consisting of seven convolutional layers, each followed by a Leaky-ReLU activation function and a Max-Pooling layer, see the details in Table~\ref{tab:encoderbigpooled}. The position of the pooling layer was one of the changes of CoRe++ compared to CoRe (in CoRe, the pooling layer was before the activation). We hypothesized that this positional change of the pooling layer could help prevent the loss of important image features. After the last convolution block, there was a flatten layer that flattened the output tensor so that it could be processed by the fully connected layer. The fully connected layer outputted the latent vector, which was a compressed representation of the 3D shape of the potato tuber. The size of the latent vector, which could be configured in our software, determined the network's ability to optimize the dataset. Intuitively, the larger the size, the higher the number of parameters in the network, meaning that it required larger datasets to properly train.

\begin{table*}[hbt!]
    \captionsetup{justification=raggedright,singlelinecheck=false} 
    \caption{Neural network architecture of the convolutional encoder of CoRe++.}
    \centering
    \begin{tabular}{c c c c c c c}
        \toprule
        \multirow{2}{*}{\textbf{Layer}} & \multirow{2}{*}{\textbf{Type}} & \multirow{2}{*}{\textbf{Kernel size}} & \multirow{2}{*}{\textbf{Strides}} & \multirow{2}{*}{\textbf{Padding}} & \multirow{2}{*}{\textbf{Activation}} & \textbf{Trainable} \\ 
         &  & & & & & \textbf{parameters} \\
         \midrule
        1 & Convolution & \(3 \times 3\) & 1 & 1 & Leaky ReLU (0.2) & 592 \\ 
        2 & Max Pooling & \(4 \times 4\) & 2 & 1 & - & 0 \\ 
        3 & Convolution & \(3 \times 3\) & 1 & 1 & LeakyReLU(0.2) & 4640 \\ 
        4 & Max Pooling & \(4 \times 4\) & 2 & 1 & - & 0 \\ 
        5 & Convolution & \(3 \times 3\) & 1 & 1 & LeakyReLU(0.2) & 18,496 \\
        6 & Max Pooling & \(4 \times 4\) & 2 & 1 & - & 0 \\
        7 & Convolution & \(3 \times 3\) & 1 & 1 & LeakyReLU(0.2) & 73,856 \\
        8 & Max Pooling & \(4 \times 4\) & 2 & 1 & - & 0 \\
        9 & Convolution & \(3 \times 3\) & 1 & 1 & LeakyReLU(0.2) & 295,168 \\
        10 & Max Pooling & \(4 \times 4\) & 2 & 1 & - & 0 \\
        11 & Convolution & \(3 \times 3\) & 1 & 1 & LeakyReLU(0.2) & 1,180,160 \\
        12 & Max Pooling & \(4 \times 4\) & 2 & 1 & - & 0 \\
        13 & Convolution & \(3 \times 3\) & 1 & 1 & LeakyReLU(0.2) & 4,719,616 \\
        14 & Max Pooling & \(4 \times 4\) & 2 & 1 & - & 0 \\
        15 & Flatten & - & - & - & - & 0 \\
        16 & Fully Connected & - & - & - & - & 131,104 \\ 
        \midrule
        \multicolumn{6}{r}{\textbf{Total:}} & \textbf{6,423,632} \\ 
        \bottomrule
    \end{tabular}
    \label{tab:encoderbigpooled}
\end{table*}

\begin{figure}[hbt!]
  \centering
    \includegraphics[width=1\linewidth]{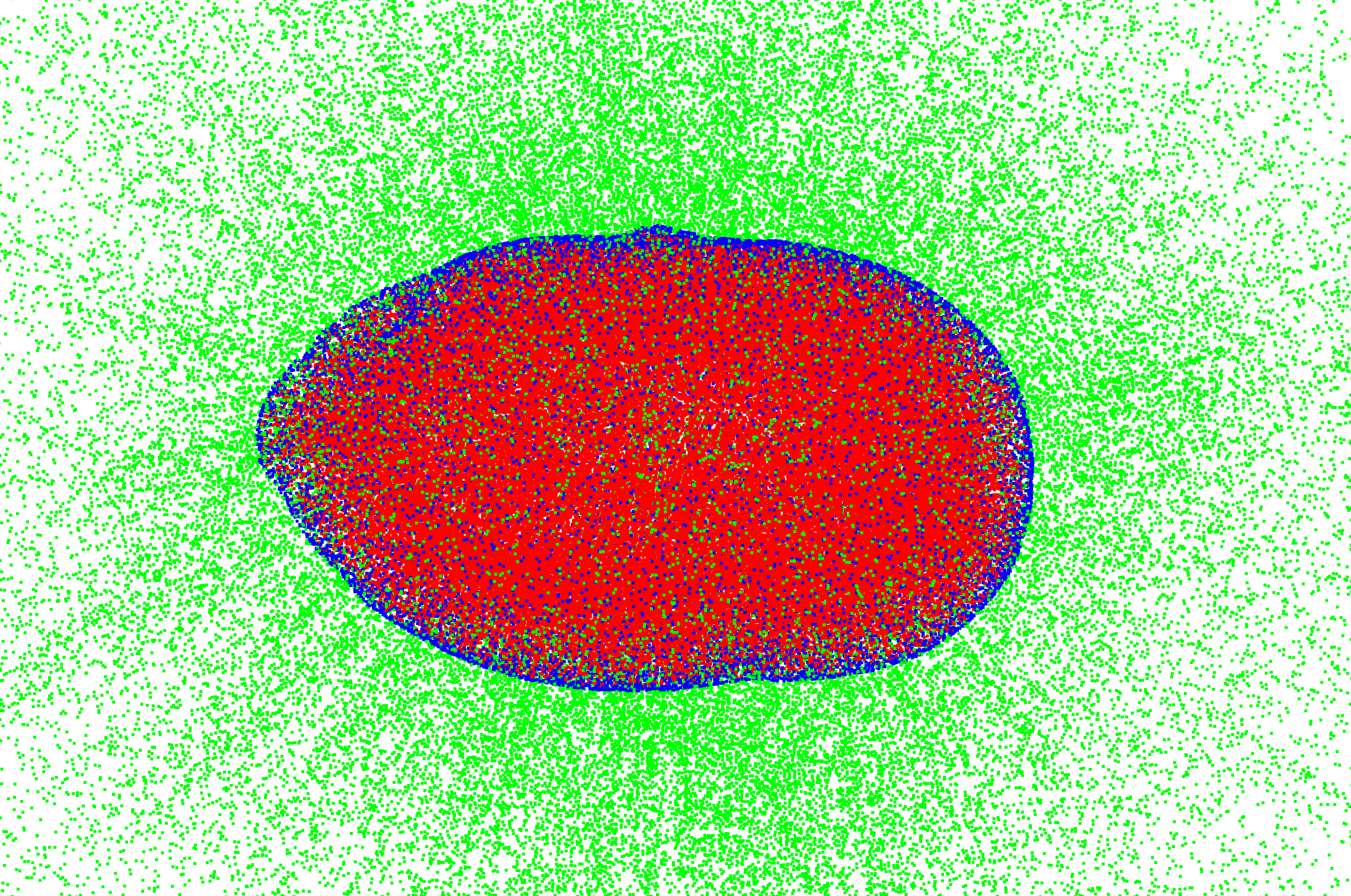}
    \caption{Visualization of the signed distances as target values for DeepSDF to learn the 3D shape of potato tubers: red points are negative (inside surface), green points are positive (outside surface), and blue points are zero (on the surface).}
    \label{fig:signed_distances}
\end{figure}

The decoder of CoRe++ was the coded-shape deep signed distance function (DeepSDF, \citealp{park2019}). This decoder was composed of nine fully connected layers with a feature dimension output of 512. The first fully connected layer had an input dimension of latent size + 3, which was the result of concatenating the latent vector with the 3-dimensional vector of the object's coordinates in x, y, z. After the last fully connected layer there was a ReLU activation function followed by a hyperbolic tangent function. The latter outputted values between -1 and 1. Conceptually, this output is the signed distance to the object's surface, in which positive values represent 3D points outside the object's surface and negative values represent 3D points inside the object's surface. The value's magnitude corresponded to the distances to the object's 3D surface, where values close to zero approximated the object's 3D shape. The concept of signed distances is visualized in Figure \ref{fig:signed_distances}.

\subsubsection{Data input, preprocessing, augmentation, and postprocessing}
\label{data_preprocessing}
The input for the CoRe++ network was a four-channel RGB-D image that was masked to the potato region and clipped to a fixed-sized image of 304 x 304 pixels (Figure \ref{fig:network}). This image dimension was chosen such that the largest tuber in our dataset would completely fit in the image. The input boxes and masks were obtained after manually annotating the potato tuber regions with the LabelMe annotation software. Note that this region clipping and masking can also be achieved with an object detection or instance segmentation model.

Two new data preprocessing techniques were added to CoRe++. The first was a depth pixel filtering algorithm, which was extracted from \citet{blok2021}, that removed all depth pixels that were further away from the depth values of the majority of the pixels in the masked depth image. This majority of pixels was assumed to represent the potato region, and helped to remove depth outliers in the depth image. The second preprocessing technique was a normalization of the remaining depth pixels between the minimum (230 mm) and maximum distance value (350 mm) between the RGB-D camera and the potato tubers on the conveyor belt. This normalization helped to increase contrast in the depth image (Figure \ref{fig:network}), which potentially leads to better model convergence. 

In addition to these new data preprocessing techniques, we also added geometric and color data augmentations to the encoder training procedure in an attempt to improve the network's generalization performance. Both geometric and color augmentations were parameterized so that the augmented images looked different but visually realistic compared to the original image. The geometric augmentations involved a random image rotation to a maximum of 45 degrees, a random horizontal flipping of the image and a random vertical flipping of the image. These augmentations mimicked different orientations of the tubers on the conveyor belt. The geometric data augmentations were applied to both the RGB image and the depth image. The color augmentations were only applied to the RGB image, and involved a random change of the brightness, saturation and hue of the image, by using the following parameters: between 0.0 and 0.5 for brightness and saturation, and between -0.1 to 0.1 for hue. These color augmentations mimicked different tuber colors and lighting conditions, helping the network to better generalize to new potato cultivars and different lighting settings.

The data augmentation for training the DeepSDF decoder involved a random scaling of the original 3D shape with scaling parameters between 0.5 and 2.0, meaning that the original shape was scaled between half of its original size and double of its size. The second data augmentation involved a rotation of the 3D shape around the z axis by a rotation value between 0.0 and 30.0 degrees. The last data augmentation was a random shear of the 3D shape of maximally 0.5 in the x direction. For each original 3D shape, 10 augmented 3D shapes were included into the train set. 

As Figure~\ref{fig:network} indicates, the output of the CoRe++ network is a completed 3D point cloud. Since it is not possible to calculate volume directly from a 3D point cloud, we added a 3D data postprocessing procedure that converted the 3D point cloud into a fully enclosed, watertight 3D mesh from which the volume could be estimated. Our 3D mesh generation consisted of selecting the predicted signed distance values less than or equal to 0.0. These values represented the 3D points on the surface and inside the potato tuber, refer to the blue- and red-colored points in Figure~\ref{fig:signed_distances}. From the selected points, a 3D convex hull shape was extracted using Open3D's GPU-accelerated hull function. The convex hull was then smoothed with a triangle-based algorithm \citep{loop1987} that divided each triangle of the hull into four smaller triangles. The value of four turned out to be optimal for smoothing the 3D shape in a high-throughput fashion. For applications that are less time-critical, a value higher than four can be chosen to make the 3D shape completion even more accurate. What followed were noise suppression procedures and an iterative voxel-based downsampling of the generated hull in case it was not watertight. The latter procedure guaranteed that each produced 3D mesh was watertight so that the volume could be calculated. 

\subsubsection{Training procedure}
\label{training_procedure}
The encoder and decoder were trained in the opposite order: first the DeepSDF decoder was trained on the complete 3D shapes in order to optimize the latent vectors. Then, the encoder was trained to fit the preprocessed RGB-D images to the target latent vectors. In Figure \ref{fig:network}, these two training procedures are visualized by the blue colored and purple colored arrows, respectively. 

At the start of the DeepSDF training, a randomly initialized latent vector was assigned to each data point. These latent vectors were then optimized along with the weights of the decoder using standard backpropagation. In our research, the DeepSDF decoder was trained for 1001 epochs with a step-based learning rate scheduler that started at a learning rate of 5$\cdot$10\textsuperscript{-4} (this value was adopted from the original DeepSDF paper). For every 300 epochs, the learning rate was halved. The decoder was trained with the Adaptive Moment Estimation (Adam) optimizer. We used the L1 loss function during training, because this loss function serves the purpose of minimizing the sum of all distance differences between the target 3D shape and the predicted 3D shape. For every 10 epochs, a weight file and a latent vector code were automatically saved. The snapshot value of 10 epochs was significantly lower than the original DeepSDF implementation (1000) and \citet{magistri2022} (500), and it allowed us to frequently identify trends of network overfitting, thereby allowing us to make a better selection of the optimal weights. The metric for selecting the optimal weights was the Chamfer distance ($d_{CD}$), which is the sum of the average closest distance from points in the ground truth 3D shape ($\mathcal{G}$) to the points in the reconstructed 3D shape ($\mathcal{S}$), and vice versa (Equation \ref{eq_cd}). The lower the Chamfer distance, the better the 3D reconstruction. The unit of the Chamfer distance was millimeter (mm). The weight file and corresponding latent vector code with the lowest Chamfer distance on the validation set were used as targets for training the encoder. 

\begin{equation}
\label{eq_cd}
d_{CD}(\mathcal{G}, \mathcal{S}) = \frac{1}{|\mathcal{G}|} \sum\limits_{x \in \mathcal{G}} \min_{y \in \mathcal{S}} \| x - y \|^2_2 + \frac{1}{|\mathcal{S}|} \sum\limits_{y \in \mathcal{S}} \min_{x \in \mathcal{G}} \| y - x \|^2_2\\
\end{equation}

\indent where $|\mathcal{G}|$ and $|\mathcal{S}|$ are the numbers of points in $\mathcal{G}$ and $\mathcal{S}$, respectively\\ 

The weights of the encoder were randomly initialized before the training started. During encoder training, the weights of the decoder were frozen. The encoder was trained for 100 epochs. The initial learning rate was 1$\cdot$10\textsuperscript{-4}, and this value was gradually decreased by an exponential learning rate scheduler of 97\%. The encoder was also trained with the Adam optimizer. We used the mean squared error loss function (MSE) to minimize the squares of the differences between the target latent vector and the predicted latent vector. We chose this loss function because it penalizes larger differences between the target and predicted latent vector more heavily than CoRe's default L1 loss function. Besides the MSE loss function, we used the contrastive loss function by \citet{magistri2022}. The rationale is that contrastive loss encourages the encoder to learn latent vectors that are well-separated in the latent space for the different potato tubers. Contrastive loss also enforces the encoder to learn latent vectors closer in the latent space for the images belonging to the same potato tuber. The contrastive loss function is summarized in Equation~\ref{eq_attrep}, where $N$ is the total number of potato tubers, $\mathbf{z}_i$ and $\mathbf{z}_j$ represent the latent vector of instances $i$ and $j$, respectively, $y_i$ and $y_j$ denote the potato tuber identifiers of instances $i$ and $j$, respectively, $\delta_{\text{rep}}$ is the margin parameter controlling the minimum separation between representations of instances with different tuber identifiers (in our research $\delta_{\text{rep}}$ was set to 0.5), $\| \cdot \|$ denotes the Euclidean distance between two latent vectors, $\| \cdot \|_+$ is the positive part of the argument, ensuring that only positive differences contributed to the loss.

\begin{equation}
\label{eq_attrep}
L_{\text{c}} = \sum_{i=1}^{N} \sum_{j=1}^{N} 
\begin{cases} 
\| \mathbf{z}_i - \mathbf{z}_j \|_{+}, & \text{if } y_i = y_j \\
\max\{0, \delta_{\text{rep}} - \| \mathbf{z}_i - \mathbf{z}_j \| \}, & \text{if } y_i \neq y_j
\end{cases} 
\end{equation}

The combined loss function for training the encoder is summarized in Equation~\ref{eq_loss}. The loss contribution values were set to 1.0 for $w_{mse}$, and 0.05 for $w_{c}$. These values were optimized in a pre-comparative experiment.

\begin{equation}
\label{eq_loss}
\mathcal{L} = w_{mse} \cdot \mathcal{L}_{mse} + w_{c} \cdot \mathcal{L}_{c}
\end{equation}

For every epoch, the encoder was inferred on the validation set with the decoder activated. The encoder weights with the lowest root mean squared error (RMSE, Equation~\ref{eq_rmse}) on the volume were stored for final evaluation on the independent test set. We chose the RMSE metric, because it was the most common metric in the scientific literature for evaluating the volumetric estimate of potato tubers. The ground truth volume ($V$) was calculated from the reconstructed 3D mesh (Section \ref{3d_reconstruction}), using the Tetrahedron method (Equation~\ref{eq_volume}). The same method was used to calculate the volume from CoRe++'s 3D reconstruction ($\hat{V}$).

\begin{equation}
\label{eq_rmse}
RMSE = \sqrt{\frac{1}{n}\sum\limits_{i=1}^{n} {\hat{V} - V}^2} 
\end{equation}
\indent where $\hat{V}$ and $V$ are the estimated volume and the ground truth volume, respectively. $n$ is the number of potato tubers in the dataset.

\begin{equation}
\label{eq_volume}
V = \sum_{i=1}^T \frac{1}{6} \left( \mathbf{v}_1 \cdot (\mathbf{v}_2 \times \mathbf{v}_3) \right)
\end{equation}
where $\mathbf{v}_1$, $\mathbf{v}_2$, $\mathbf{v}_3$ are the three vertices representing a triangle in the mesh. $T$ is the total number of triangles in the mesh.

\subsection{Experiments}
\label{experiments}
\subsubsection{The effect of the latent size on the 3D shape completion result}
\label{effect_latent_size}
For 3D shape completion, it is important that the encoder-compressed latent vector contains high-level distinguishable 3D features that generalize well on new and untrained shapes. Because the latent vector plays such an important role in an encoder-decoder network like ours, we set up an experiment that tested what was the optimal latent vector size for completing the 3D shape of potato tubers. We investigated six sizes: 8, 16, 32, 64, 128, and 256. Latent size 32 was used in the original study by \citet{magistri2022}. Our hypothesis was that the smaller sizes, such as 8 and 16, compressed the data too much, possibly losing critical details, leading to suboptimal generalization. The larger sizes, such as 128 and 256, potentially captured more noise, leading to overfitting and suboptimal shape generalization. The medium sizes, 32 and 64, probably struck a better balance between compression and generalizability because they allowed the model to efficiently capture key features.

To better understand the optimal size of the latent vector, we conducted an additional experiment with our CoRe++ network. For each of the six tested latent sizes, we interpolated the values of the latent vector of the smallest potato tuber and the largest in the test set. Interpolating between these two latent vectors can reveal how well the network has learned to represent size variations. A smooth transition in the generated 3D shapes and sizes can indicate that the latent space is well-structured, meaning that the network can effectively generate realistic 3D shapes for unseen inputs, i.e., generalize better \citep{mi2021}.

The experiment was tested with 3 networks: the DeepSDF decoder architecture without the encoder, the original CoRe network of \citet{magistri2022}, and our CoRe++ network. The DeepSDF decoder-only architecture was tested to better understand where the largest effect of latent size was: in the encoder or in the decoder. Testing the original CoRe network of \citet{magistri2022} provided insight into a potential trend commonly shared between CoRe and CoRe++. It also gave a more detailed overview on the potential improvement of CoRe++ over CoRe. 

\begin{figure}[hbt!]
  \centering
    \includegraphics[width=1\linewidth]{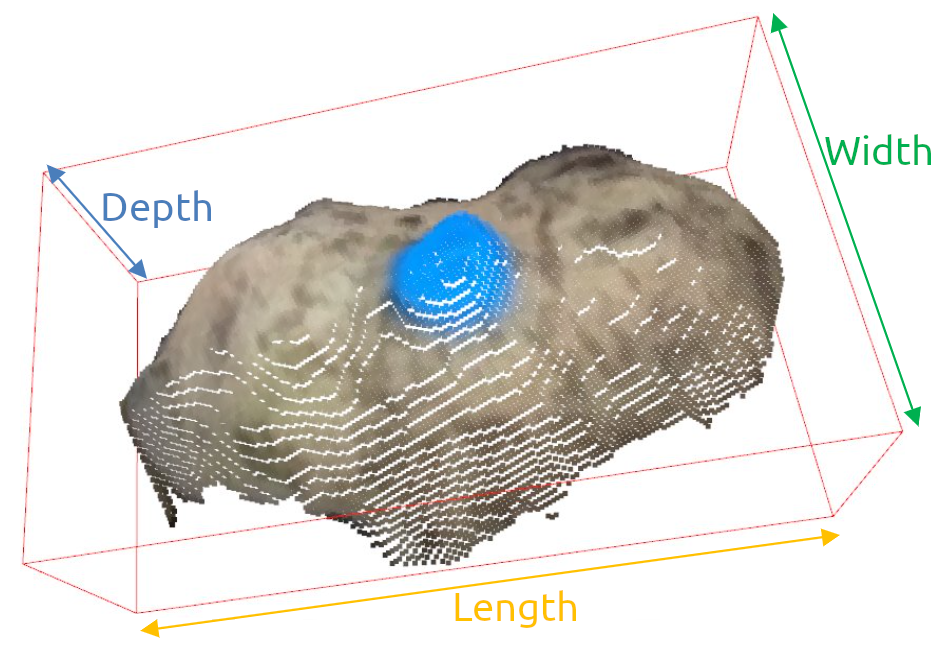}
    \caption{To determine whether the accuracy requirement was met, we trained a linear regression model on the length, width and depth estimates of the oriented 3D bounding box generated from the partially completed point cloud.}
    \label{fig:input_regression}
\end{figure}

Our evaluation was based on two requirements: accuracy and analysis speed. The accuracy requirement was met if the RMSE on the volume (Equation~\ref{eq_rmse}) was lower than that of a standard linear regression model. Input to that linear regression model were the length, width and depth estimates of the partially completed point cloud (Figure~\ref{fig:input_regression}). This point cloud was obtained by converting the filtered RGB-D image (Figure \ref{fig:network}) into 3D points using the intrinsic camera parameters of the RGB-D camera in Open3D software. From the generated point cloud, the oriented 3D bounding box was obtained from which the length, width and depth dimensions were extracted (Figure~\ref{fig:input_regression}) as input parameters for training the linear regression model. The linear regression model was trained on the same train and validation set and tested on the same independent test set (Section \ref{dataset_splits}). Our trained linear regression model resulted a RMSE of 31.1 ml on the volumetric estimate of the 1425 test images. As such, if the 3D shape completion method had a RMSE lower than 31.1 ml, then the accuracy requirement was met. 

The analysis speed requirement was met if the 3D shape completion was finished in less than 16 milliseconds (ms) per potato tuber. This maximum analysis time was calculated based on the highest number of potato tubers present in a single image in our dataset: 59. The average throughput time of the potato tubers on the harvester was 0.95 seconds, and this value was derived by dividing the average number of 28.5 frames per tuber (9658/339, Section \ref{dataset_splits}) by the camera acquisition rate of 30 frames per second. With the maximum number of tubers in a single image and the average throughput time, we estimated that the 3D shape completion method should analyze up to 62 potato tubers per second (59/0.95). This equaled to 16 ms per tuber (1000/62).

\subsubsection{The effect of the potato size, potato cultivar, and image analysis region on the 3D shape completion result}

The goal of this experiment was to gather insights into the performance of our CoRe++ shape completion method when applied in the field. The analysis was conducted with the best performing latent size found in the previous experiment. Our first practical analysis was on the effect of the tuber size on the 3D shape completion result. This analysis is important because it provides information about the degree of generalizability of our method in a particular field. Under typical field conditions, a wide variety of tuber sizes move over the harvester's conveyor belt, and this size variability can influence yield estimation. Our method must handle this variability effectively, ensuring accurate volumetric estimates regardless of tuber size. Such flexibility is crucial for making reliable yield predictions in real-world harvest situations. We decided to perform the analysis on four different size classes, which were extracted from our dataset distribution (Figure \ref{fig:histogram}): small tubers with a volume between 0 and 100 ml, small to medium tubers with a volume between 100 and 150 ml, medium to large tubers with a volume between 150 and 200 ml, and large tubers with a volume between 200 and 500 ml. The ranges were chosen so that a relatively balanced number of tubers ended up in the four classes. 

The second practical analysis was on the effect of the potato cultivar on the 3D shape completion result. This analysis provided insights into the degree of generalizability of the 3D completion method at the farm level, as the majority of farmers grow multiple cultivars in a season. We wanted to test if our method can provide accurate volumetric estimates regardless of the potato cultivar. We conducted the experiment with three potato cultivars (Corolle, Kitahime and Sayaka) that differed in shape and size. Corolle was the cultivar with the most elongated and smallest tubers. Kitahime was the cultivar with the most spherical and medium-sized tubers. Sayaka was the cultivar with the largest tubers and the greatest diversity in shape. This may be due to the fact that Sayaka was the cultivar with the most samples in our dataset. As shape metrics, we calculated the elongation factor and the concavity factor. The elongation factor was obtained by dividing the longest dimension of the 3D bounding box of the ground truth mesh by the shortest dimension. Spherical shapes have an elongation factor close to 1 and elongated shapes have an elongation factor closer to 2. The concavity factor was obtained by calculating the Chamfer distance between the original ground truth 3D mesh and the convex hull shape of that mesh. Higher Chamfer distances mean that there are more valleys or concave regions on the surface of the potato tuber. 

Our final practical analysis focused on identifying the horizontal image region where ideally the 3D shape completion should be performed. The reason for this analysis is that it is neither computationally desirable nor necessary to analyze all 20 to 30 RGB-D frames of each tuber when it moves over the conveyor belt. The computational efficiency can be improved by performing the 3D shape completion only in the image region with the average lowest RMSE for volumetric estimation. In our analysis, we examined thirteen horizontal image regions (Table~\ref{tab:pixel_regions}).

\begin{table}[hbt!]
    \captionsetup{justification=raggedright,singlelinecheck=false} 
    \caption{Summary of the investigated horizontal image regions.}
    \centering
    \begin{tabular}{l c}
        \toprule
        Region & Horizontal image region \\ 
        & [pixels] \\
        \midrule
        1 & 0 - 100\\
        2 & 100 - 150\\
        3 & 150 - 200\\
        4 & 200 - 250\\
        5 & 250 - 300\\
        6 & 300 - 350\\
        7 & 350 - 400\\
        8 & 400 - 450\\
        9 & 450 - 500\\
        10 & 500 - 550\\
        11 & 550 - 600\\
        12 & 600 - 650\\
        13 & 650 - 720\\ 
        \bottomrule
    \end{tabular}
    \label{tab:pixel_regions}
\end{table}

\subsubsection{Ablation studies}
Two ablation studies were performed as a final experiment. The first ablation study was conducted on the additions to CoRe++, which helped us to better understand the impact of the individual additions to the overall performance. The second ablation study was conducted on the components that were commonly shared between CoRe++ and CoRe. This ablation study helped us to better understand the impact of changes in the input data and network architecture of the convolutional encoder. 

The first ablation study consisted of examining the impact of seven additions made to CoRe++. The first two examined ablations consisted of individually deactivating the depth normalization and the depth filtering when training the convolutional encoder (both data preprocessing steps are described in Section \ref{data_preprocessing}). After that, we examined the impact of deactivating the data augmentation when training the convolutional encoder. The fourth ablation examined the effect of repositioning the Max-Pooling layer before the LeakyReLU activation, as was the case in the original CoRe implementation (refer to Section \ref{core_architecture}). The fifth examined ablation was changing the MSE loss function back to the L1 loss function, as was the case in CoRe (refer to Section \ref{training_procedure}). The sixth examined ablation consisted of replacing CoRe++'s method for automatically determining the best network weights (using GPU-based 3D mesh generation and the RMSE volume metric) with CoRe's original method (using marching cubes \citep{lorensen1987} mesh generation and Chamfer distance metric). The seventh examined ablation consisted of deactivating CoRe++'s 3D smoothing technique which was described in Section \ref{training_procedure}. 

The second ablation study consisted of examining the effects of ten ablations in the input data and network architecture that were commonly shared between CoRe++ and CoRe. The first ablation involved training CoRe++ with a single-channel depth image instead of the original four-channel RGB-D image. This ablation gave us a better understanding of the effect of adding color channels to the input data for completing the 3D shape. The second ablation involved training CoRe++ with an RGB-D image clipped to the bounding box instead of the original mask. This ablation gave us insight into how the final application on the potato harvester should look like: one based on an object detection model or one based on an instance segmentation model. Logically, this ablation also gave insight into the future annotation effort, which would be significantly higher when using an instance segmentation model. The third ablation involved simplifying the encoder's network architecture from seven to five convolutional blocks (layers 1-10 in Table \ref{tab:encoderbigpooled} followed by a flatten layer and a fully connected layer). This ablation gave us insight into the effect on performance when using an even lighter network in the situation of hardware constraints. The fourth ablation involved removing the pooling layers (the even layers in Table \ref{tab:encoderbigpooled}) to investigate their effect on the overall performance. The fifth ablation involved replacing CoRe++'s original LeakyReLU activation with the standard ReLU activation. The sixth ablation involved disabling the contrastive loss element in the loss function so that only the MSE loss was used during training. The seventh to the tenth ablation involved increasing or decreasing the learning rate in increments of two and five. The resulting learning rates were 5$\cdot$10\textsuperscript{-4} (learning rate $\cdot$ 5), 2$\cdot$10\textsuperscript{-4} (learning rate $\cdot$ 2), 5$\cdot$10\textsuperscript{-5} (learning rate $\cdot$ 0.5) and 2$\cdot$10\textsuperscript{-5} (learning rate $\cdot$ 0.2).

\subsection{Evaluation}
\label{evaluation}
For evaluating the 3D point cloud completion in our experiments, we used the Chamfer distance (Equation \ref{eq_cd}, Section \ref{training_procedure}) as performance metric. Besides the Chamfer distance, we also calculated the precision, recall and the F-score. The precision ($p$) was the percentage of reconstructed points within a certain distance ($d$) to the ground truth point cloud (Equation \ref{eq_p}). As such, it represented the accuracy of the 3D shape completion. The recall ($r$) was the percentage of ground truth points within a certain distance ($d$) to the reconstructed point cloud (Equation \ref{eq_r}). The recall represented the completeness of the 3D shape completion. For our evaluation, we used a distance ($d$) threshold  of 5.0 mm. This threshold was adopted from \citet{magistri2022}, so that we could compare our results one-on-one with their results. After calculating the precision and recall, the F-score was obtained (Equation \ref{eq_f1}). The F-score was the harmonic mean between the precision and the recall, and it represented the percentage of 3D points that were correctly reconstructed.

\begin{equation}
\label{eq_p}
precision(d) = \frac{100}{|\mathcal{S}|} \sum\limits_{y \in \mathcal{S}} \Bigr\llbracket \min_{x \in \mathcal{G}} \| y - x \| < d \Bigr\rrbracket\\
\end{equation}

\begin{equation}
\label{eq_r}
recall(d) = \frac{100}{|\mathcal{G}|} \sum\limits_{x \in \mathcal{G}} \Bigr\llbracket \min_{y \in \mathcal{S}} \| x - y \| < d \Bigr\rrbracket\\
\end{equation}

\begin{equation}
\label{eq_f1}
\textit{f-score}(d) = \frac{2 \cdot p(d) \cdot r(d)}{p(d) + r(d)}\\
\end{equation}

For evaluating the volumetric estimate in our experiments, we used the RMSE metric, as presented in Equation~\ref{eq_rmse}. To evaluate the processing speed, we calculated the average 3D shape completion time on the test images. This processing speed analysis was performed on a Lenovo Legion Pro 7 16IRX8H laptop with an Intel i9-13900HX 2.2 GHz CPU with 32GB RAM and a NVIDIA GeForce RTX 4090 Laptop GPU with 16GB memory. This laptop was considered a suitable laptop for use on an operational harvester.

\section{Results}
\label{results}
\subsection{The effect of the latent size on the 3D shape completion result}
Table~\ref{tab:latent_size} summarizes the results of the 3D shape completion for the three 3D completion methods and the six latent sizes. Interestingly, the effect of the latent size was marginal when testing the DeepSDF decoder-only network. For this network, the smallest latent size of 8 resulted in the lowest Chamfer distance and RMSE on the volume. One possible explanation for this result is that the DeepSDF decoder uses an iterative optimization process during inference to extract the best possible latent vector. This iterative process probably helped to optimize the latent vector for each potato shape in the test set, making the 3D shape completion slow (33 seconds on average), but also better optimized for each of the six tested latent sizes. 

\begin{figure*}[hbt!]
  \centering
    \includegraphics[width=1\textwidth]{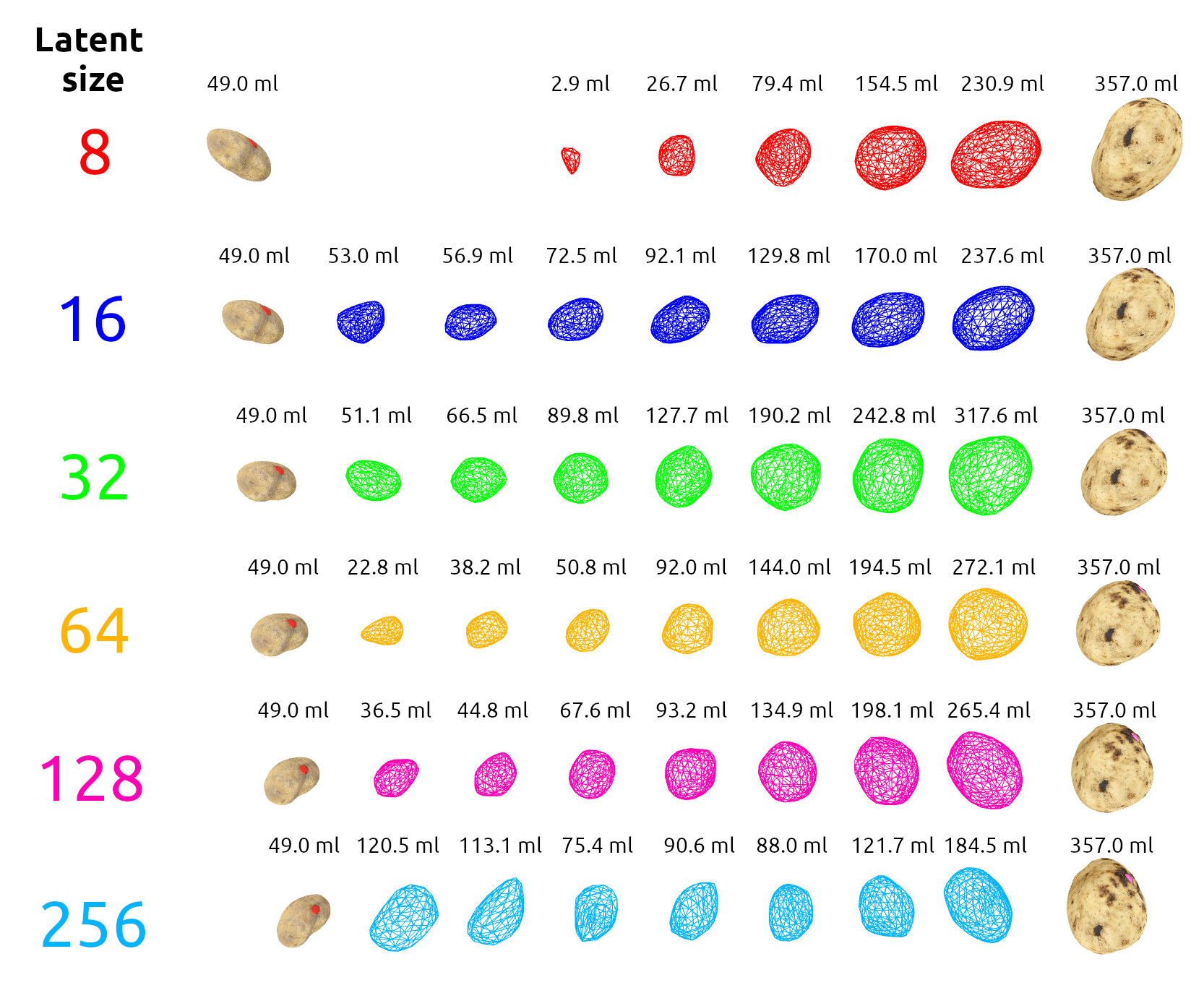}
    \caption{3D shape completion results for the seven latent space interpolations, visualized for each of the six tested latent sizes. The far left and far right meshes are the smallest and largest tubers in the test set between which the interpolation was performed.}
    \label{fig:latent_space}
\end{figure*}

For both CoRe and CoRe++, the best latent size was 32, followed by 64. This result is consistent with our hypothesis that the medium-sized latent vectors strike a better balance between compression and generalizability. Figure~\ref{fig:latent_space} shows that latent sizes 32 and 64 have the most diverse 3D shapes and sizes after latent space interpolation. Compared to the other sizes, latent sizes 32 and 64 have both spherical and elongated 3D shapes, and the sizes show a proper sequential build-up, indicating a better generalized latent space. Latent size 8 and 256 are both worse in 3D shape completion performance (Table~\ref{tab:latent_size}) and latent space interpolation (Figure~\ref{fig:latent_space}). Latent size 8 could only produce five valid 3D shapes out of the seven interpolated latent vectors, while latent size 256 failed to produce a sequential size build-up. Latent sizes 16 and 128 produced mainly elongated 3D shapes, meaning that they had a limited ability to produce the more spherical shapes.  

In terms of 3D shape completion and volumetric estimate, CoRe++ outperformed CoRe for five of the six latent sizes: 16, 32, 64, 128, 256 (Table~\ref{tab:latent_size}). The difference between the 3D shape completion result of CoRe++ and CoRe is visualized in Figure~\ref{fig:reconstruction_results}. As for CoRe++, the reconstructed 3D shapes are smooth and they approximate the real shape of the potato tuber. With CoRe, the 3D shapes are rougher and spikier, leading to less resemblance to the real potato shape, larger Chamfer distances and larger volumetric errors.

Only for CoRe++ and latent sizes 32 and 64, the accuracy requirement was met as the corresponding RMSE values were less than 31.1 ml (this was the baseline value when using linear regression). The requirement for analysis speed was met for each latent size for both CoRe and CoRe++, as all of the analysis times were less than 16 ms (Table~\ref{tab:latent_size}). Regarding analysis time, no clear trend was observable when increasing or decreasing the latent vector size, implying that it does not matter for the analysis speed which size is chosen. 

\subsection{The effect of the potato size, potato cultivar, and image analysis region on the 3D shape completion result}

This experiment was conducted with CoRe++ and latent size 32, as this was the best performing combination in the previous experiment. Table~\ref{tab:results_size} summarizes the effect of the potato size on the 3D shape completion result. There is a trend that the RMSE on the volumetric estimate is larger when the tuber is larger. A possible explanation is that the larger tubers are more concave (Table~\ref{tab:results_size}), meaning that they have more valleys and variations in their 3D curvature, making it harder to accurately reconstruct the 3D shape (especially when the concave parts are facing downward with respect to the camera's perspective). It was demonstrated that especially the DeepSDF decoder produced large volumetric errors on tubers with a high concavity factor. This indicates that DeepSDF struggled in reconstructing the concave parts. The encoder produced the largest volumetric errors on the largest tubers in the test set (their volume was in between 225 and 357 ml), and this may be due to the under-representation of these tubers in our dataset (Figure~\ref{fig:histogram}). In-depth analysis on the five potato tubers with the largest 3D shape completion errors confirmed these trends, as these tubers had a high concavity factor (up to 0.45), were abnormally large or both. CoRe++'s largest error of 104 ml (which is visualized in Figure~\ref{fig:bad_reconstruction}), was on the second largest and the second most concave potato tuber in the test set. Thus, for a better generalization performance on the potato harvester, it may be beneficial to increase data diversity and expand the train set with more examples of large, concave tubers. 

Table~\ref{tab:results_cultivar} summarizes the effect of the potato cultivar, and thus indirectly the shape of the potato tuber, on the 3D shape completion results. There is a trend that the Chamfer distance is lower when the tubers have a more spherical shape (i.e. elongation factors closer to 1.0). For the RMSE on the volumetric estimate, there is no clear trend regarding the tuber shape. What may explain this result is that the different cultivars had different tuber sizes: the cultivar Corolle had the relative smallest tubers and it was already shown in Table~\ref{tab:results_size} that the RMSE is lowest when the tubers are smaller. The largest RMSE was observed on Sayaka, which was the cultivar with the largest tubers on average, but it was also the cultivar with the largest number of samples, meaning that there was probably more diversity in tuber shape and size. Thus, for better generalization performance on cultivars with highly varied tuber characteristics, it may be beneficial to extend the data augmentations when training the encoder and decoder.

\begin{figure*}[hbt!]
  \centering
    \includegraphics[width=0.95\textwidth]{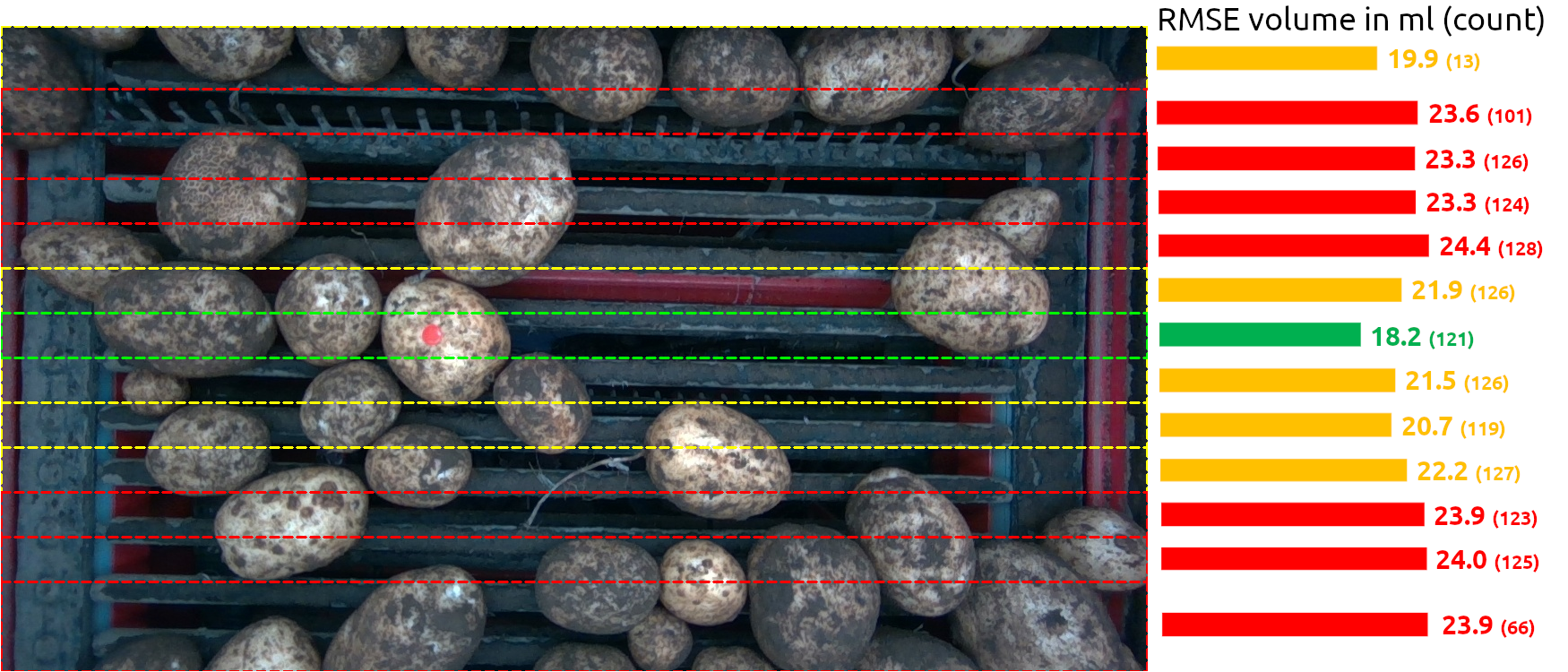}
    \caption{Root mean square errors (RMSE) visualized for thirteen horizontal image regions. The green-colored region in the center of the image between 350 and 400 pixels had the lowest RMSE of 18.2 ml.}
    \label{fig:performance_regions}
\end{figure*}

Figure~\ref{fig:performance_regions} visualizes the RMSE on the volumetric estimate for the thirteen different image regions. The smallest RMSE of 18.2 ml was observed in the central horizontal region of the image between 350 and 400 pixels. By performing the 3D shape completion only in this image region, the computational efficiency of the volumetric estimates can be improved in a practical setup. The largest RMSE values were observed in the lower and upper parts of the image, indicating that these regions are not recommended for performing the 3D shape completion. These results may also indicate that some form of lens distortion has occurred in the peripheral regions of the camera's field of view, leading to poorer 3D shape completion in these regions. There may also have been a higher degree of occlusion in these regions.

\subsection{Ablation studies}
\begin{figure*}[hbt!]
  \centering
  \subfloat[] {\includegraphics[width=1\textwidth]{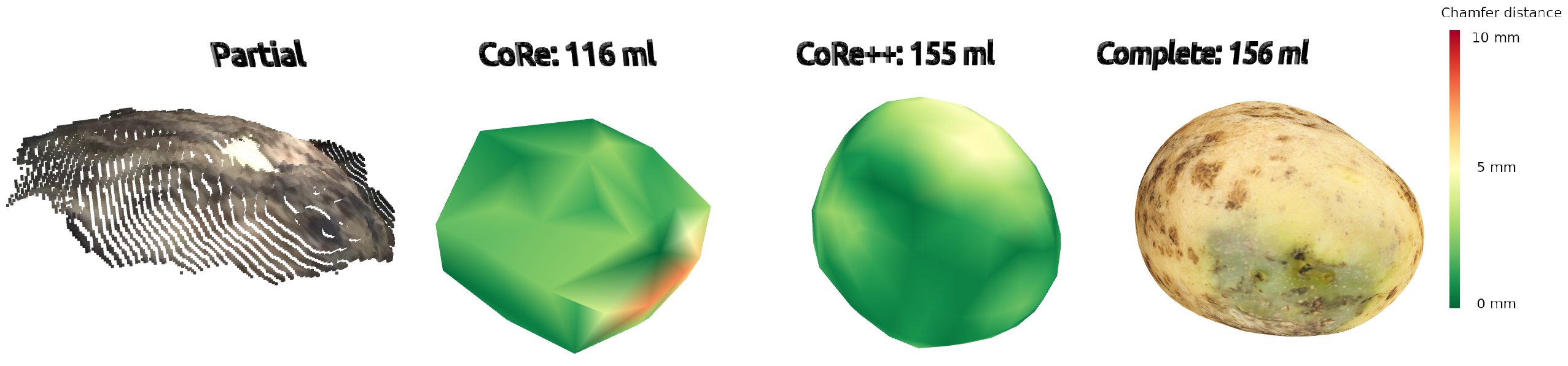}\label{fig:good_reconstruction}}
  \hfill
  \subfloat[] {\includegraphics[width=1\textwidth]{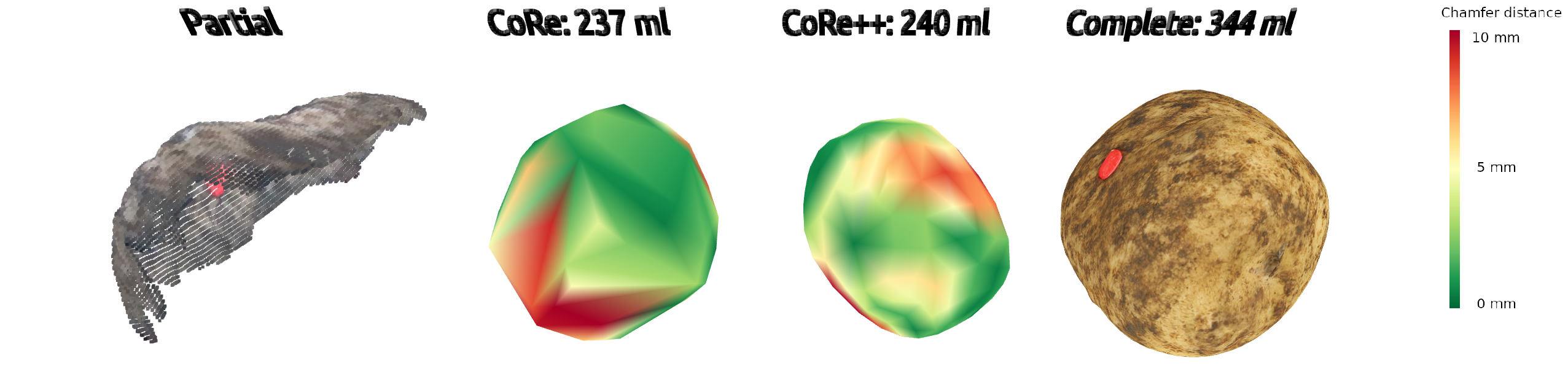}\label{fig:bad_reconstruction}}
  \caption{(a) CoRe++'s best 3D shape completion result was achieved on a medium-sized potato tuber. (b) CoRe++'s worst 3D completion result was achieved on a large-sized and irregularly-shaped potato tuber. In (a) and (b), the partial point cloud is visualized on the left, the completion result from CoRe is visualized on the second from the left, and the one from CoRe++ is visualized on the second from the right. Most right is the 3D ground truth shape.}
  \label{fig:reconstruction_results}
\end{figure*}

The ablation study on CoRe++'s additions (Table~\ref{tab:ablation_study}) shows that the largest contribution to the overall performance was made by CoRe++'s validation method, which was based on GPU mesh generation and RMSE validation metric. Changing this validation method back to the original validation method of CoRe with marching cubes mesh generation and Chamfer distance validation metric, led to the largest increase in both Chamfer distance (+32.1\%) and RMSE on volumetric estimate (+122.1\%). An obvious explanation for this result is that the final volumetric estimate benefits from having the RMSE optimized during training. Another explanation is that CoRe's original marching cubes method seems unable to accurately reconstruct the 3D shapes of the potato tubers. This may be due to the chosen grid density of the marching cubes method which was optimized for high-throughput 3D reconstruction, but possibly came at the expense of the accuracy (something that can also be observed in Figure~\ref{fig:reconstruction_results}). Three other additions to CoRe++ that significantly improved the overall performance were the two data preprocessing steps (highlighted by the first and second ablation) and the loss function modification to MSE (highlighted by the fifth ablation).

Regarding the general ablation study (Table~\ref{tab:ablation_study}), there are a few interesting outcomes. First, the choice of the activation function had a large impact on the overall performance. The fifth ablation shows that after replacing the LeakyReLU activation with a standard ReLU, the Chamfer distance increased by 46.4\% and the RMSE by 96.0\%. This outcome is consistent with that of \citet{tomar2022}, who also found that ReLU under-performs compared to the more advanced variants of ReLU when testing an autoencoder on the Global Wheat Head dataset \citep{david2020}. In our case, we think that the LeakyReLU activation resulted in a better and faster network convergence during training, due to LeakyReLU's ability to maintain non-zero gradients for negative inputs. A second interesting outcome was the relatively large impact of the RGB color channels on the overall performance. Our initial expectation was that only the depth image would contain relevant features for completing the 3D shape. There are two possible explanations for this result. First, the depth image has typically more noise than the RGB image, which may lead to less good feature extraction when only using the depth image. Second, the data augmentations for the RGB color channels were more extensive than for the depth channel, which may have resulted in a better generalization performance when using RGB-D images rather than just depth images. A third interesting outcome is that training without constrastive loss resulted in a lower Chamfer distance and a higher F-score. One explanation for this result is that the contrastive loss may only be beneficial if the images were obtained from different camera perspectives, as was the case in \citet{magistri2022}. In our experiment, the potato tubers were photographed from the same top-view camera perspective, which resulted in relatively similar 3D shapes, making it more difficult for the contrastive loss function to separate the embedding space. Two other observations from the ablation study that are useful for implementing CoRe++ on an operational harvester are: an instance segmentation algorithm is preferred over an object detection algorithm (as highlighted by the second ablation), and a lighter encoder network may be more favorable for high-throughput shape completion, but it comes at the expense of the accuracy (as highlighted by the third ablation). In summary, the two ablation studies highlight that CoRe++'s validation method, data preprocessing, LeakyReLU activation, and inclusion of RGB channels were key to improving the 3D shape completion of potato tubers.

\begin{table*}[hbt!]
\caption{The effect of the latent size on the 3D shape completion results with DeepSDF, CoRe, and CoRe++. The upward arrows indicate the higher the better, and the downward arrows the lower the better. The values in bold are the best performing values per 3D shape completion method. Please refer to Section \ref{effect_latent_size} for the accuracy and speed requirements and Section~\ref{evaluation} for the performance metrics.}
\begin{center}
 \begin{tabular}{l c c c c c c c c c} 
 \toprule
 3D shape completion & latent & $d_{CD}$ & f-score & precision & recall & RMSE & time & & \\
  method & size & [mm] $\downarrow$ & [\%] $\uparrow$ & [\%] $\uparrow$ & [\%] $\uparrow$ & [ml] $\downarrow$ & [ms] $\downarrow$ & acc. & speed\\
 \midrule
  {\multirow{6}{3.5cm}{\raggedright DeepSDF \\ \citep{park2019}}} & 8 & \textbf{1.5} & 98.1 & 97.9 & 98.4 & \textbf{7.2} & 32636.5 & \cmark & \xmark\\
  & 16 & 1.7 & 98.8 & 98.8 & 98.7 & 11.1 & 32669.0 & \cmark & \xmark\\
  & 32 & 1.8 & 99.2 & 99.2 & 99.2 & 7.5 & 32580.9 & \cmark & \xmark\\
 & 64 & 1.8 & 98.8 & 98.8 & 98.8 & 16.8 &  32792.3 & \cmark & \xmark\\
  & 128 & 1.8 & \textbf{99.3} & \textbf{99.4} & \textbf{99.3} & 13.3 & 32919.1 & \cmark & \xmark\\
  & 256 & 1.9 & 97.3 & 97.4 & 97.3 & 13.4 & 33003.6 & \cmark & \xmark\\
 \midrule
  \multirow{6}{3.5cm}{\raggedright CoRe \\ \citep{magistri2022}} & 8 & 6.0 & 50.8 & 53.2 & 49.7 & 60.6 & 8.5 & \xmark & \cmark\\
  & 16 & 3.7 & 71.8 & 73.8 & 70.0 & 50.3 & 9.3 & \xmark & \cmark\\
  & 32 & \textbf{3.1} & \textbf{81.4} & \textbf{81.5} & \textbf{81.5} & \textbf{36.9} & 7.5 & \xmark & \cmark\\
  & 64 & 3.4 & 76.1 & 74.1 & 78.6 & 41.6 & 8.3 & \xmark & \cmark\\
  & 128 & 3.3 & 78.3 & 78.7 & 78.0 & 43.9 & 8.0 & \xmark & \cmark\\
  & 256 & 5.3 & 58.3 & 58.8 & 58.6 & 90.0 & \textbf{6.7} & \xmark & \cmark\\
  \midrule
  \multirow{6}{3.5cm}{\raggedright CoRe++ \\ (ours)} & 8 & 7.9 & 38.1 & 39.8 & 37.6 & 68.8 & \textbf{9.1} & \xmark & \cmark\\
  & 16 & 4.2 & 65.4 & 66.8 & 64.4 & 44.6 & 13.0 & \xmark & \cmark\\
  & 32 & \textbf{2.8} & \textbf{85.0} & \textbf{85.2} & \textbf{85.0} & \textbf{22.6} & 9.9 & \cmark & \cmark\\
  & 64 & 2.9 & 83.2 & 83.8 & 82.8 & 28.1 & 9.4 & \cmark & \cmark\\
  & 128 & 3.3 & 78.6 & 81.1 & 76.4 & 35.7 & 9.3 & \xmark & \cmark\\
  & 256 & 4.5 & 62.5 & 63.7 & 61.8 & 65.0 & 10.2 & \xmark & \cmark\\
 \bottomrule
 \end{tabular}
 \end{center}
 \label{tab:latent_size}
\end{table*}

\begin{table*}[hbt!]
    \captionsetup{justification=raggedright,singlelinecheck=false} 
    \caption{3D shape completion results expressed for the four size classes. Count summarizes the total number of RGB-D frames analyzed per size class.}
    \centering
    \begin{tabular}{l c c c c c c c c c}
        \toprule
        Volume & Count & Elongation & Concavity & $d_{CD}$ & f-score & precision & recall & RMSE & rel. error \\
        \text{[ml]} & & factor & factor [mm] & [mm] $\downarrow$ & [\%] $\uparrow$ & [\%] $\uparrow$ & [\%] $\uparrow$ & [ml] $\downarrow$ & [\%] $\downarrow$ \\ \midrule
         0-100 & 361 & 1.6 & 0.2 & 2.5 & 89.3 & 89.0 & 89.8 & 16.8 & 19.1\\
         100-150 & 364 & 1.4 & 0.3 & 3.0 & 82.3 & 82.1 & 82.6 & 18.2 & 12.5\\
         150-200 & 277 & 1.4 & 0.3 & 2.7 & 85.4 & 85.8 & 85.1 & 21.7 & 10.0\\
         200-500 & 423 & 1.5 & 0.4 & 2.9 & 83.5 & 84.2 & 82.9 & 29.7 & 9.0\\ 
         \bottomrule
    \end{tabular}
    \label{tab:results_size}
\end{table*}

\begin{table*}[hbt!]
    \captionsetup{justification=raggedright,singlelinecheck=false} 
    \caption{3D shape completion results expressed for the three tested potato cultivars. Count summarizes the total number of RGB-D frames analyzed per cultivar.}
    \centering
    \begin{tabular}{l c c c c c c c c c}
        \toprule
        Potato & Count & Elongation & Concavity & $d_{CD}$ & f-score & precision & recall & RMSE & rel. error\\
        cultivar & & factor & factor [mm] & [mm] $\downarrow$ & [\%] $\uparrow$ & [\%] $\uparrow$ & [\%] $\uparrow$ & [ml] $\downarrow$ & [\%] $\downarrow$\\ 
        \midrule
         Corolle & 291 & 1.9 & 0.3 & 2.9 & 83.9 & 84.1 & 83.8 & 17.6 & 15.4 \\
         Sayaka & 869 & 1.5 & 0.3 & 2.8 & 84.5 & 84.9 & 84.2 & 24.8 & 11.8\\ 
         Kitahime & 265 & 1.2 & 0.3 & 2.5 & 88.0 & 87.2 & 88.9 & 19.3 & 12.6\\ 
         \bottomrule
    \end{tabular}
    \label{tab:results_cultivar}
\end{table*}

\clearpage
\begin{table*}[hbt!]
\caption{Performance metrics for the two ablation studies relative to the best performance of CoRe++.}
\centering
\begin{tabular}{l l c c c c c c c c c}
    \toprule
    & & \multicolumn{2}{c}{$d_{CD} (mm)$ $\downarrow$} & & \multicolumn{2}{c}{f-score $\uparrow$} & & \multicolumn{2}{c}{RMSE (ml) $\downarrow$} &\\
    \cline{3-4} \cline{6-7} \cline{9-10}
    Ablation & Category & abs & rel & & abs & rel & & abs & rel &\\ 
    \midrule
    CoRe++ & Baseline & 2.8 & - & & 85.0 & - & & 22.6 & - &\\ 
    \midrule
    \multicolumn{11}{c}{\textit{\textbf{Ablation study on CoRe++'s additions}}}\\
    \midrule
    No depth normalization & Data preprocessing & 3.3 & +17.9\% & & 78.6 & -7.5\% & & 34.3 & +51.8\% &\\
    No depth filtering & Data preprocessing & 3.4 & +21.4\% & & 75.9 & -10.7\% & & 41.1 & +81.9\% & \\
    No data augmentation & Data augmentation & 2.9 & +3.6\% & & 83.9 & -1.3\% & & 32.4 & +43.4\% &\\
    $\text{Act-Pool} \rightarrow \text{Pool-Act}$ & Network changes & 2.9 & +3.6\% & & 83.2 & -2.1\% & & 31.0 & +37.2\% & \\
    $\text{MSE loss} \rightarrow \text{L1 loss}$ & Loss function & 3.1 & +10.7\% & & 81.3 & -4.4\% & & 31.4 & +38.9\% &\\
    $\text{Val} \rightarrow \text{CoRe val}$  & Train validation & 3.7 & \textbf{+32.1\%} & & 72.0 & \textbf{-15.3\%} & & 50.2 & \textbf{+122.1\%} & \\
    $\text{Smoothing} \rightarrow \text{Custom}$ & 3D postprocessing & 2.8 & 0.0\% & & 85.1 & +0.2\% & & 24.9 & +10.2\% & \\
    \midrule
    \multicolumn{11}{c}{\textit{\textbf{General ablation study}}}\\\hline
    $\text{RGB-D} \rightarrow \text{D}$ & Data preprocessing & 3.5 & +25.0\% & & 75.4 & -11.3\% & & 40.5 & +79.2\% &\\
    $\text{Mask} \rightarrow \text{Box}$ & Data preprocessing & 3.2 & +14.3\% & & 78.9 & -7.2\% & & 33.0 & +46.0\% &\\
    $\text{7} \rightarrow \text{5}$ Conv. blocks & Network changes & 3.0 & +7.1\% & & 82.3 & -3.2\% & & 32.0 & +41.6\% & \\
    No pooling layers & Network changes & 2.9 & +3.6\% & & 83.0 & -2.4\% & & 29.7 & +31.4\% &\\
    $\text{LeakyReLU} \rightarrow \text{ReLU}$ & Network changes & 4.1 & \textbf{+46.4\%} & & 68.1 & \textbf{-19.1\%} & & 44.3 & \textbf{+96.0\%} & \\
    No contrastive loss & Loss function & 2.7 & -3.6\% & & 86.3 & +1.5\% & & 23.6 & +4.4\% &\\
    $\text{LR} \rightarrow \text{LR$\cdot$5}$  & Learning rate & 2.9 & +3.6\% & & 83.6 & -1.6\% & & 32.1 & +42.0\% &\\
    $\text{LR} \rightarrow \text{LR$\cdot$2}$ & Learning rate & 2.8 & 0.0\% & & 85.2 & +0.2\% & & 28.8 & +27.4\% &\\
    $\text{LR} \rightarrow \text{LR$\cdot$0.5}$ & Learning rate & 3.0 & +7.1\% & & 81.9 & -3.6\% &  & 31.0 & +37.2\% &\\
    $\text{LR} \rightarrow \text{LR$\cdot$0.2}$ & Learning rate & 3.3 & +17.9\% & & 77.5 & -8.8\% & & 39.9 & +76.5\% &\\
    \bottomrule
\end{tabular}
\label{tab:ablation_study}
\end{table*}

\clearpage
\section{Discussion}
\label{discussion}
Of the six latent sizes, we observed that a latent size of 32 outperformed the other sizes in terms of 3D shape completion and volumetric estimate. This suggests that a moderate latent size strikes a balance between representational capacity and model complexity, allowing RGB-D images to be encoded more effectively while not overfitting. The latter was clearly demonstrated by the latent space interpolation, which generated the most realistic potato shapes and sizes for latent size 32. Compared to the literature, our observations are similar to those of \citet{ahmed2022}, who found that a latent size of 28 was best for optimizing a convolutional variational autoencoder on spectral topographic maps. Since \citet{ahmed2022} did their research on 25 latent sizes, this suggests that even better results could have been obtained if we had tested more latent sizes. In future work, a finer-grained latent size analysis could help identify optimal latent configurations for 3D shape completion tasks in agriculture, thereby enabling better estimates of crop shapes, volumes, and total yield.

With CoRe++ better 3D shape completion results were obtained compared to the linear regression model and the original CoRe implementation of \citet{magistri2022}. When we compare our results with those of \citet{magistri2022}, we can conclude that our results on potato are similar (in comparison with strawberry), or better (in comparison with sweet pepper). An important remark is that the obtained results may depend on the average complexity of the shape that has to be completed. Shape complexity can impact model generalizability, affecting its ability to adapt to unseen crop instances. Potato has on average a less complex shape than sweet pepper, making it easier for the network to learn a generic shape enabling a better generalization performance. In future research, we want to test our CoRe++ model on crops or fruits with a more complex shape, such as pineapple, dragon fruit and Romanesco broccoli. We encourage fellow researchers to test our publicly available software on other 3D shapes within the agricultural domain or beyond. 

Our research has provided valuable insights and steps towards the practical application of CoRe++ on a potato harvester. Nevertheless, we think there are potential improvements in software and hardware that could further improve the performance. Regarding software, we would like to explore the use of other 3D shape completion networks, such as the one by \citet{magistri2024}, who used a transformer network. The obtained results with this new transformer network on sweet pepper and strawberry were significantly better than those of the original CoRe implementation \citep{magistri2022}. An advantage of using a transformer network is that it is end-to-end trainable, enabling simultaneous optimization of all network weights, which is advantageous over our CoRe++ network that needs to be trained in two stages. A hardware improvement that could potentially improve the 3D shape completion is equipping the conveyor belt with rotating rollers in the area where the camera is placed. The rotating rollers cause the potato tubers to rotate gradually as they pass underneath the camera, meaning that almost the entire shape of the potato can be photographed. This can both simplify and improve the 3D shape completion. In a scenario like this, it would also be useful to test whether the contrastive loss function has an improving effect on completing the 3D shapes of the rotating potato tubers. 

To further test the applicability of CoRe++ on a potato harvester, it is important to conduct a more in-depth evaluation on potato tubers of different cultivars. This evaluation will provide a better understanding of the overall generalization performance of CoRe++ at the farm level. In such an analysis, it is important to perform the test on potato tubers without the colored thumbtack, because the thumbtack’s color and shape may have influenced CoRe++'s extraction of distinguishable features during training. This could have unintentionally caused a stronger model trigger towards colors and shapes that do not appear in practical situations. A future evaluation could benefit from performing the analysis on a more balanced dataset. Our in-depth analysis revealed that the largest errors were on small and large tubers and those with irregular shape, and this might be problematic when analyzing an entire field. Future research should therefore focus on obtaining more balanced datasets or developing training tools that help the network better deal with under-represented samples.

When applying CoRe++ in practice, it is important to investigate its performance, scalability and interpretability as part of a larger system. In integrated systems, the performance of CoRe++ will be affected by the instance segmentation algorithm that provides the binary mask. It is therefore important to investigate the sensitivity of CoRe++ to the segmentation output of such instance segmentation algorithm. In larger integrated systems, there may be constraints towards the scalability of CoRe++. For instance, when using multi-row harvesters, there will be a larger throughput of potato tubers on the conveyor belt and this puts pressure on CoRe++'s ability to complete the 3D shape of all tubers on time. A straightforward solution is increasing the computational resources, but this will make the overall system more expensive. We believe it is better to develop a multi-threaded software approach, in which all 3D shape completion tasks are distributed across multiple concurrently-running threads. Regarding the system's output, it is important to consider that the results must be post-processed in such a way that they can be easily interpreted by the end user. In its current setup, CoRe++ produces a volumetric estimation for each potato tuber, but these estimates have to be spatially summarized in such a way that a farmer can easily interpret the output. This means that in an integrated system, the outputs of CoRe++ have to be geo-referenced by means of a global navigation satellite system (GNSS). Such a GNSS-integrated system provides opportunities for integration with in-season crop growth data from drones, so that additional insights into the impact of management decisions can be obtained.

\section{Conclusions}
\label{conclusions}
In this study, we investigated a high-throughput 3D shape completion network for its suitability for estimating the volume of potato tubers on an operational harvester. Our research revealed that latent size 32 had the best 3D shape completion result. With that latent size, our CoRe++ network had an RMSE of 22.6 ml on the volumetric estimate, and this was better than the RMSE of the linear regression (31.1 ml) and the original CoRe network (36.9 ml). We also found that the RMSE of CoRe++ could be further reduced to 18.2 ml when performing the 3D shape completion in the center of the RGB-D image. With an average analysis time of 10 milliseconds per potato tuber, CoRe++ enables a high-throughput 3D shape completion up to 100 potato tubers per second. Hence, we can conclude that our network is able to quickly and accurately estimate the volume of fast-moving potato tubers on an operational harvester. Our method provides real-time potato tuber yield estimates, giving farmers valuable insights to further optimize planting, fertilization, and harvest strategies in subsequent growing seasons. This can help to increase yield, reduce waste, and better meet market demands\textemdash effects not easily achieved by current analytics. The continual progress in 3D shape completion and machine learning algorithms will enable more accurate and more useful yield monitoring, as well as mitigating the effects of occlusion and partial visibility in robotic harvesting, quality control and sorting.

\section*{Acknowledgements}
We would like to thank Okada Farm for providing the potato field on which we acquired the RGB-D images. We thank Ting Jiang, Sylvain Grison, Yuto Imachi and Irena Drofová for their help with the image acquisition and ground truth measurements. 

\printcredits

\bibliographystyle{bib-style}

\bibliography{refs}

\end{document}